\documentclass[11pt]{article}

\usepackage[margin=1in]{geometry}
\usepackage[T1]{fontenc}
\usepackage[utf8]{inputenc}
\usepackage{amsmath, amssymb, amsfonts, amsthm}
\usepackage{mathabx}
\usepackage{graphicx}
\usepackage{hyperref}
\usepackage{booktabs}
\usepackage{array}
\usepackage{multicol}
\usepackage{algorithm}
\usepackage{algpseudocode}
\usepackage{listings}
\usepackage{xcolor}
\usepackage{natbib}
\usepackage{bbm}
\usepackage{bm}
\usepackage{tikz}
\usetikzlibrary{decorations.pathreplacing}
\usepackage{stmaryrd}
\usepackage{enumitem}
\usepackage{threeparttable}
\usepackage{makecell}
\usepackage{pifont}
\usepackage{float}

\usepackage{tabularx}

\theoremstyle{definition}
\newtheorem{definition}{Definition}
\newtheorem{example}{Example}

\def\imp{\rightarrow}

\def\sltn{\texttt{sLTN}}
\def\ltn{\texttt{LTN}}
\def\I{\mathcal{I}}                 

\def\Sig{\Sigma}                    
\def\S{\mathcal{S}}                 
\def\D{\mathcal{D}}                 
\def\X{\mathcal{X}}                 
\def\C{\mathcal{C}}                 
\def\F{\mathcal{F}}                 
\def\Pr{\mathcal{P}}                
\def\SV{\Xi}  
\def\SR{\mathcal{R}}  
\def\sort{\mathrm{sort}}
\def\dims{\mathrm{dims}}
\newcommand{\bnfalt}{\;\Big|\;}
\def\indim{\mathrm{dims_{in}}}
\def\outdim{\mathrm{dims_{out}}}

\def\outsort{\mathrm{sort_{out}}}

\def\axes{\mathrm{axes}}
\def\fv{\mathrm{fv}}
\def\fsv{\mathrm{fsv}}

\DeclareMathOperator*{\argmax}{argmax}

\DeclareMathOperator*{\SatAgg}{SatAgg}

\newcommand{\code}[1]{\texttt{#1}}
\newcommand{\dvar}[1]{\bar\alpha(#1)}

\newcommand{\val}[1]{v(#1)}

\title{\sltn: Structural Logic Tensor Networks}

\author{
  Davide Rinaldi \\ Nokia Bell Labs \\ davide.rinaldi@nokia-bell-labs.com
  \and
  Luciano Serafini \\ Fondazione Bruno Kessler \\ serafini@fbk.eu
}
\date{\today}

\begin{document}

\maketitle

\begin{abstract}
Logic Tensor Networks (LTN) provide a neurosymbolic framework in which first-order logic is interpreted through tensor operations, enabling logical constraints to be integrated with differentiable learning. However, the original formulation of LTN is primarily suited to data represented as flat collections of individuals, and does not explicitly capture structural organization such as temporal order, sequential position, or graph connectivity.

We introduce sLTN, an extension of LTN that makes structural dimensions first-class elements of the language. Structural dimensions represent named tensor axes associated with domain-specific organization, such as time steps, sequence positions, or graph nodes. They can be quantified explicitly, related through structural relations, and used to express temporal, sequential, and relational constraints directly at the logical level.

We formalize the syntax and fuzzy tensor semantics of sLTN and show that, in the absence of structural dimensions, the framework recovers the original LTN semantics as a special case. We further describe a PyTorch implementation based on a declarative signature, formula parsing, and tensorial interpretation. The framework is illustrated on representative temporal and sequential reasoning examples. This paper serves as a companion to the \sltn{} library, available at \url{https://github.com/logictensornetworks/sltn}.
\end{abstract}

\section{Introduction}
\label{s:intro}

Logic Tensor Networks (\ltn{}) \citep{badreddine_logic_2022} is a neurosymbolic framework that integrates first-order logic with neural networks. \ltn{} is based on first-order logic where non-logical symbols, such as constants, variables, functions, and predicates, are grounded in tensors or differentiable functions over tensors. Logical connectives and quantifiers are interpreted using fuzzy semantics, selected from a range of fuzzy logics. This interpretation makes it possible to compute the value of every term and the (fuzzy) truth value of every formula in the language.

The main feature of \ltn{} is that every term and formula is interpreted in a differentiable interpretation, parameterized by learnable functions, whose parameters can be optimized through gradient descent with the objective of maximizing the overall truth value of a set of logical formulas provided.
This allows us to rephrase ``learning'' in terms of ``maximum satisfiability, '' unifying ``logical reasoning'' and ``machine learning'' in a unique framework.

The paper \cite{serafini2016learning} introduced, for the first time, the core ideas of \ltn{} together with a first implementation based on \code{TensorFlow 1}. \cite{badreddine_logic_2022} consolidated the initial framework and extended it with key features such as domains, guarded quantification, and diagonalized variables. It also presented a set of basic examples demonstrating the potential of the framework. This version migrated the implementation to \code{TensorFlow 2}. Later, a \code{PyTorch} implementation was developed and described in \cite{carraro2024ltntorch}.
Since its first version, \ltn{} has been successfully applied to a variety of tasks involving both learning from data and logical reasoning. One of its most common application areas is semantic image interpretation \cite{LTNIJCAI,%
donadello2019compensating,%
bouaziz2025enhancing,%
manigrasso2026boosting,%
bergamin2025integrating,%
colamonaco2026weakly}. In these works, \ltn{} is primarily used to impose ontological constraints on the classes and relations between entities appearing in images. They consistently show that injecting logical knowledge during training improves performance compared with purely data-driven supervision.
\ltn{} has also been applied to recommendation systems \cite{carraro2023overcoming,carraro2024mitigating,eckert2026extending,tan2024ontomedrec}. In this setting, logical formulas expressing general commonsense recommendation rules are used to compensate for data sparsity, particularly for new users and items. Besides improving predictive performance, the use of logical knowledge also provides a mechanism for explaining the recommendations.
Furthermore, \ltn{} has been used to develop decision support systems in medicine \cite{mondal2025logic} and computational biology \cite{gao2025enhancing}. In \cite{umili2023grounding,de2026neuro,gaikwad12026neuro}, \ltn{} is employed to specify requirements on dynamic processes with the objective of training models to classify correct process executions. Finally, \cite{dai2025large,manigrasso2024probing,haufe2026large,yang2025neuro} combine large language models (LLMs) with \ltn{} to compare and integrate LLM-based reasoning with logical reasoning.
Several extensions of the original \ltn{} have also been proposed. For instance, \cite{boscarato2026first} extends \ltn{} with temporal modal operators for linear temporal logic; \cite{djenouri2024neurosymbolic} integrates \ltn{} with Vision Transformers; \cite{gang2026feedltn} extends \ltn{} to support federated learning while preserving data privacy; and \cite{upreti2026logic} integrates \ltn{} into a generative adversarial network architecture.
Finally, \cite{de2024enhancing} extends the family of fuzzy logics available in \ltn{} with additional fuzzy operators based on uninorms. This extension increases both the flexibility of the framework and its performance on a range of neurosymbolic learning tasks.

Despite its effectiveness across these applications, the original \ltn framework is primarily designed for settings in which data are modeled as flat collections of individuals. This assumption becomes restrictive when data are inherently organized along temporal, sequential, or relational axes, such as time steps in a time series, tokens in a sequence, or nodes in a graph. In such settings, structural organization is part of the semantics of the problem itself rather than a mere implementation detail. Reasoning may need to refer explicitly to positional or relational patterns, for example to express temporal persistence, neighborhood constraints in graphs, or dependencies between ordered positions. 

To address this limitation, we introduce Structural Logic Tensor Networks \sltn{}, an extension of \ltn{} designed to support logical reasoning over structured data. The central idea of \sltn{} is to make structural organization an explicit component of the language. In particular, \sltn{} introduces \emph{structural dimensions} as first-class entities representing positional or relational axes such as time steps, graph nodes, or sequence positions. As a result, logical expressions can refer not only to ordinary first-order objects but also to the structural dimensions along which those objects are organized. A second key notion is that of \emph{structural relations}, interpreted as Boolean or fuzzy masks over structural dimensions. For example, over a temporal dimension 
$t$
one may define a structural relation $\mathrm{next}(t,t')$ expressing adjacency between consecutive time steps, and formulate constraints such as
\begin{align}
\forall t,t' \mid \mathrm{next}(t,t') : \bigl(A(x_t) \rightarrow A(x_{t'})\bigr),
\label{eq:Ax_t->Ax_next_t}
\end{align}
Formula~\eqref{eq:Ax_t->Ax_next_t}
states that whenever a property holds at one time step, it should also hold at the next. More generally, structural relations allow the logical language to refer directly to organization encoded by temporal order, graph adjacency, or other structured dependencies.

As in the original \ltn{}, the framework supports differentiable fuzzy semantics for logical connectives and quantifiers. \sltn{} extends these mechanisms to formulas involving structural dimensions and structural relations, thereby enabling logical reasoning over structured inputs while preserving compatibility with gradient-based optimization.

From an implementation perspective, \sltn{} separates syntax from semantics. Symbols are first declared in a \code{Signature}, formulas are parsed and validated against that signature, and tensorial meanings are assigned only subsequently through an \code{Interpretation}. Evaluation is then performed in PyTorch by means of named-axis tensor operations.

The contributions of this paper are threefold. First, we extend the language of \ltn{} with structural dimensions and structural formulas, making structured organization an explicit object of logical modelling. Second, we provide a semantics that integrates these extensions with differentiable fuzzy connectives and quantifier aggregation. Third, we describe a modular implementation of the resulting framework in Python and PyTorch. The full implementation of \sltn{}, together with example notebooks, is available at \url{https://github.com/logictensornetworks/sltn}.

The remainder of the paper is organized as follows:
Section~\ref{s:sltn-logic} introduces the syntax and semantics of \sltn{}, including terms and formulas with structural variables and their tensor-based denotation;
Section~\ref{s:learning} formulates learning in \sltn{} as the gradient-based optimization of differentiable satisfiability objectives.
Section~\ref{s:impl} presents the implementation details of the \sltn{} library;
Section~\ref{s:relwork} discusses related work;
Finally, Section~\ref{s:concl} concludes the paper and outlines directions for future work.
\section{The language of \sltn{}}
\label{s:sltn-logic}

In this section, we introduce the formal language of \sltn{}.
The language extends a standard many-sorted first-order signature
with \emph{structural dimensions}, \emph{structural variables}, and
\emph{structural relations}. 

Structural dimensions provide a syntactic 
way to refer to positions inside structured objects, 
such as time steps in a sequence, or nodes in a graph. 
Semantically, they determine the tensor axes along which
expressions are aligned, selected, masked, and aggregated. Accordingly,
\sltn{} distinguishes between first-order variables, which range over
individuals of a declared sort, and structural variables, which range over
indices of a declared structural dimension.

The formal presentation of \sltn{} closely follows its implementation. We first
define signatures, then assign tensor-level interpretations to their symbols,
and finally extend these interpretations compositionally to terms and formulas.
Alongside the formal definitions, we use a simple video-classification example
to illustrate the corresponding concrete \sltn{} syntax. The example is meant
primarily as a tutorial device.

\subsection{Signature}
\label{s:signature}

An \sltn{} signature is a tuple
\[
\Sig=(\S,\D,\C,\X,\F,\Pr,\SV,\SR,\sort{},\dims{}),
\]
of sets of symbols for 
sorts ($\S$), dimensions ($\D$), constants ($\C$), individual variables ($\X$), functions ($\F$), 
predicates ($\Pr$), structural variables $(\SV)$, and structural relations ($\SR$). The maps \(\sort\) and
\(\dims\) specify the sorts and structural dimensions of these symbols. Specifically: 
\[
\sort : \C\cup\X \to \S,
\qquad
\sort_{\mathrm{in}} :
\F\cup\Pr \to \S^{*},
\qquad
\outsort :
\F \to \S .
\]
Thus, constants and variables have a declared sort; functions and predicates
have a finite sequence of input sorts; and functions additionally have an
output sort. The sorts \code{Real} and \code{Bool} are included in every 
signature by default.

The structural typing of constants, variables, functions, and predicates is
specified by
\[
\dims : \C\cup\X \to \D^{*},
\qquad
\dims_{\mathrm{in}} :
\F\cup\Pr \to \D^{*},
\qquad
\outdim :
\F\cup\Pr \to \D^{*}.
\]
For \(a\in\C\cup\X\), \(\dims(a)\) is the possibly empty tuple of structural dimensions carried by occurrences of \(a\). For \(g\in\F\cup\Pr\), \(\indim(g)\) is the dimensional profile consumed by an application of \(g\), whereas \(\outdim(g)\) is the structural profile produced by the resulting term or formula. Importantly, \(\indim(g)\) is associated with the symbol \(g\) as a whole, not with each argument separately. Hence the dimensions consumed by \(g\) may be distributed across its arguments: different arguments may carry different subsets of \(\indim(g)\). This convention is made precise in
Section~\ref{s:semantics}.

Structural variables and structural relations are typed by dimension-specific maps:
\[
\mathit{dim}:\SV\to\D,
\qquad
\dims_{\mathrm{rel}}:\SR\to\D^*.
\]
If \(\mathit{dim}(\alpha)=d\), then the structural variable \(\alpha\) ranges
over indices of the structural dimension \(d\). If
\[
\dims_{\mathrm{rel}}(R)=\langle d_1,\ldots,d_n\rangle,
\]
then \(R\) is a relation over tuples of structural indices of dimensions
\(d_1,\ldots,d_n\). 
The same structural dimension may occur more than once in
\(\dims_{\mathrm{rel}}(R)\).



We use the following video-classification task as a running example \sltn{}.

\begin{figure}[h]
\centering
\includegraphics[width=.8\linewidth]{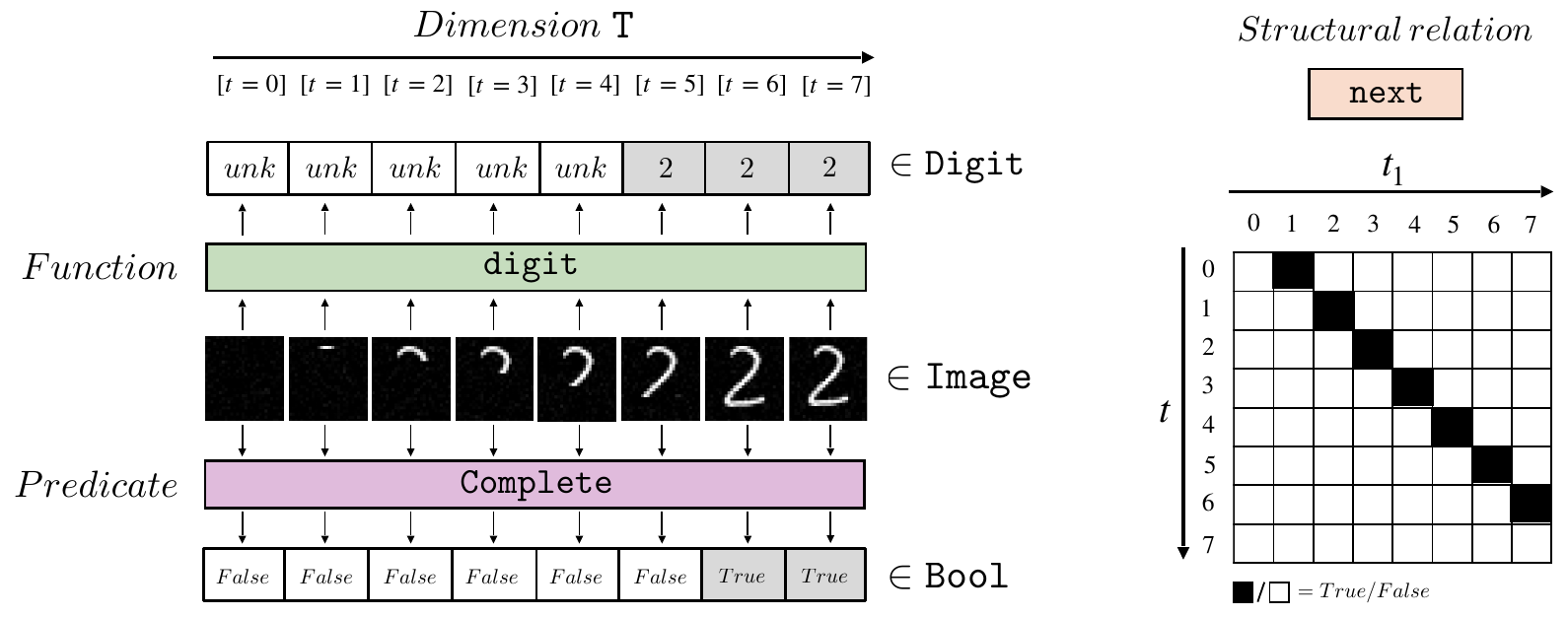}
\caption{Running video-classification example: a sequence of frames shows a
handwritten digit being progressively drawn. Function \(\mathrm{Digit}\) outputs a digit class at each time step, while predicate \(\mathrm{Complete}\) outputs a truth value. Structural relation \(\mathrm{next}\) encodes pairs of consecutive time indices.}
\label{fig:running_example}
\end{figure}

\begin{example}[Running example]
\label{ex:video-signature}

We consider videos generated from MNIST digits. Each sample is a sequence \(x\) of \(T\) frames. There are two types of videos. In an \emph{appear} video, a single digit is revealed monotonically: the frames are blank before a randomly chosen start time, the digit is gradually revealed, and it remains fully visible after completion. In a \emph{non-appear} video, the initial reveal is followed by one or more hide--reveal cycles, so that the visibility trace is generally non-monotone. Each video has a label \(y\), represented as a one-hot vector over eleven classes: the ten digit classes and an additional class
\(\mathtt{unknown}\). For appear videos, \(y\) is the label of the drawn digit. For non-appear videos, \(y\) is \(\mathtt{unknown}\), even if the digit is visible in some frames.

The signature contains two sorts,
\(\mathtt{Image}\) and \(\mathtt{Digit}\), and one structural dimension \(\mathtt{T}\) for time. The variable \(x\) denotes a video and carries the dimension \(\mathtt{T}\). The variable \(y\) denotes a per-video label and has no structural dimension. The function \(\mathtt{digit}\) maps a single image to a digit prediction. The predicate \(\mathtt{Complete}\) maps a single frame to a truth value indicating whether the digit is complete in that frame. The predicate \(\mathtt{appear}\) makes a whole-video judgement: it consumes the temporal axis but does not produce any. Finally, \(\mathtt{unknown}\) is the constant denoting the pseudo-class for incomplete or non-appearing videos, and \(\mathtt{next}\) relates consecutive time indices. 

In \sltn{}, the signature is declared as follows:
\begin{lstlisting}[language=Python]
sig = Signature("video")

sig.sort("Image")
sig.sort("Digit")

sig.dimension("T")
sig.structural_variable("t", "T")

sig.constant("unknown", "Digit")

sig.variable("x", "Image", dims=["T"])
sig.variable("y", "Digit")

sig.function("digit", ["Image"], "Digit")
sig.predicate("Complete", ["Image"])
sig.predicate("appear", ["Image", "Digit"], input_dims=["T"], output_dims=[])
sig.predicate("=d", ["Digit", "Digit"], infix=True)

sig.structural_relation("next", ["T", "T"])
\end{lstlisting}

This example illustrates different structural behaviours. The function \(\mathtt{digit}\) and the predicate \(\mathtt{Complete}\) have neither input nor output structural dimensions: they are applied to individual frames and may
therefore be evaluated pointwise along an existing temporal axis. By contrast, the predicate \(\mathtt{appear}\) consumes the temporal axis and produces no structural output, thereby expressing a whole-sequence judgement. Figure~\ref{fig:running_example} provides a visual representation.
\end{example}

The declarations in a signature determine the admissible vocabulary of a theory: they specify which symbols are available, together with their sorts and dimensional profiles. The resulting sort and dimension constraints are used below to define which terms and formulas are well formed.

\subsection{Terms and Formulas}
\label{s:grammar}

Terms and formulas are generated by the following grammar:
\renewcommand{\arraystretch}{1.8}
\[
\begin{array}{rr@{\ }l}
  \textit{term} & t ::= & x \bnfalt c \bnfalt f(t,\dots,t)
    \bnfalt t[\alpha_1,\dots,\alpha_k] \bnfalt t[\alpha=n] \\
\textit{formula} & \phi,\psi ::= & p(t,\dots,t)
     \bnfalt \lnot\phi \bnfalt \phi \land \psi \bnfalt \phi \lor \psi
     \bnfalt \phi \imp \psi \bnfalt \phi \leftrightarrow \psi \\
   && \bnfalt Q\, x_1,\dots,x_h :\, \phi 
       \bnfalt Q\, x_1,\dots,x_h \mid \psi :\, \phi \\
   && \bnfalt Q\, (x_1,\dots,x_h) :\, \phi  
      \bnfalt Q\, (x_1,\dots,x_h) \mid \psi :\, \phi \\
   && \bnfalt Q\, \alpha_1,\dots,\alpha_k\ :\, \phi
   \bnfalt Q\,  \alpha_1,\dots,\alpha_k \mid \rho :\, \phi \\
   && \bnfalt \phi[\alpha_1,\dots,\alpha_k]
      \bnfalt \phi[\alpha=n] \\
\textit{structural\, formula} & \rho,\sigma ::= & R(\alpha_1,\dots,\alpha_n)
     \bnfalt \lnot \rho \bnfalt \rho \land \sigma \bnfalt \rho \lor \sigma
     \bnfalt \rho \imp \sigma \bnfalt \rho \leftrightarrow \sigma .
\end{array}
\]
The symbol \bnfalt is a meta-level separator between grammar alternatives and must not be confused with the symbol \(\mid\), which is part of the object language and introduces guards. Here:
\begin{itemize}[itemsep=1pt]
    \item \(x,x_i\in\X\), \(c\in\C\), \(f\in\F\), \(p\in\Pr\),
    \(R\in\SR\), \(d\in\D\), \(\alpha,\alpha_i\in\SV\) and \(n,k,h\in\mathbb{N}\).

    \item \(Q\in\{\forall,\exists\}\) ranges over ordinary first-order
    quantifiers. When \(Q\) binds first-order variables
    \(x_i\), it denotes ordinary first-order quantification. When \(Q\) binds structural variables \(\alpha_i\), it denotes structural quantification over indices of the corresponding declared dimensions.

    \item The form \(Q\,x_1,\dots,x_h:\phi\) denotes ordinary first-order
    quantification over the listed variables, whereas
    \(Q\,(x_1,\dots,x_h):\phi\) denotes \emph{diagonal quantification}, that
    is, quantification over aligned tuples of the listed variables rather than
    over their full Cartesian product.

    \item Quantifiers may be guarded. The form
    \(Q\,x_1,\dots,x_h \mid \psi : \phi\) evaluates \(\phi\) on assignments
    of \(x_1,\dots,x_h\) satisfying the guard \(\psi\). Analogously,
    \(Q\,\alpha_1,\dots,\alpha_n \mid \rho : \phi\) evaluates \(\phi\) on
    assignments of structural indices satisfying the structural formula \(\rho\).

    \item The annotations \(t[\alpha_1,\dots,\alpha_k]\) and
    \(\phi[\alpha_1,\dots,\alpha_k]\) rename the output structural axes of a
    term or formula. In the current \sltn{} implementation, renaming is
    all-or-nothing: if the expression has \(k\) output structural axes, they
    are renamed according to the ordered list
    \((\alpha_1,\dots,\alpha_k)\). This is useful when the same structural
    dimension is used under different aliases in a formula.

    \item The annotations \(t[\alpha=n]\) and \(\phi[\alpha=n]\) select the slice at index \(n\) along the structural axis named by \(\alpha\).
\end{itemize}

In addition to being generated by the grammar, expressions must satisfy the typing constraints induced by the signature. Function and predicate applications must match the declared input sorts and structural profiles; equality-like predicates must compare terms of compatible sorts and structural profiles; axis annotations and selections must refer to structural variables of the appropriate dimensions; and binders must bind variables consistently with their declared sorts or dimensions. Expressions satisfying these constraints are called \emph{well formed}.

\paragraph{Syntax conventions in \sltn{}.}
The grammar above defines the abstract syntax of the \sltn{} language. The
concrete \sltn{} implementation follows this grammar, but also provides a few
syntactic conveniences that are resolved before interpretation:
\begin{enumerate}[itemsep=1pt]
\item Structural quantifiers may be written directly over a dimension name. Internally, an implicit structural variable is created, with the same name as the corresponding dimension.
\item If \(\alpha\) is declared over dimension \(d\), then indexed or primed
variants such as \(\alpha_1\), \(\alpha_2\), \(\alpha1\), \(\alpha2\), and \(\alpha'\) are treated as distinct structural variables ranging over \(d\). This convention is useful when several indices of the same structural dimension occur in one formula, as in \(\mathrm{next}(t,t_1)\) in Example~\ref{ex:video-signature}.

\item Explicit axis annotations may be omitted when they are uniquely inferable from the signature and surrounding binders. Nevertheless, explicit annotations are recommended when they improve readability or disambiguate repeated structural dimensions.

\item Logical connectives and quantifiers may be written using either symbolic
or textual notation, as summarized in Table~\ref{tab:concrete-connectives}.
The keywords \code{forall}, \code{exists}, \code{not}, \code{and},
\code{or}, \code{implies}, and \code{iff}, as well as
\(\forall,\exists,\neg,\sim,\wedge,\&,\vee,\rightarrow,\leftrightarrow,=,:,\) and \(\mid\), are reserved by the parser and cannot be used as signature symbols.
\end{enumerate}

\begin{table}[h]
\centering
{\renewcommand{\arraystretch}{1.2}
\begin{tabular}{lcc}
\toprule
Construct & \makecell{Mathematical \\ notation} & \makecell{Concrete syntax \\ alternatives} \\
\midrule
Negation
& \(\lnot \phi\)
& $\sim$\code{phi}, \code{not phi}, \code{¬phi} \\

Conjunction
& \(\phi \land \psi\)
& \code{phi \& psi}, \code{phi and psi}, \code{phi }$\wedge$\code{ psi} \\

Disjunction
& \(\phi \lor \psi\)
& \code{phi or psi}, \code{phi }$\vee$\code{ psi} \\

Implication
& \(\phi \imp \psi\)
& \code{phi -> psi}, \code{phi implies psi}, \code{phi → psi} \\

Equivalence
& \(\phi \leftrightarrow \psi\)
& \code{phi <-> psi}, \code{phi iff psi}, \code{phi }$\leftrightarrow$\code{ psi} \\

Universal quantification
& \(\forall x:\phi\)
& \code{forall x:phi}, $\forall$\code{ x:phi} \\

Existential quantification
& \(\exists x:\phi\)
& \code{exists x:phi}, $\exists$\code{ x:phi} \\

Guarded quantification
& \(Q\; x \mid\psi:\phi\)
& \code{Q x |psi:phi}\\

Structural quantification
& \(Q\; \alpha \mid \rho:\phi\)
& \code{Q \(\alpha\)| rho:phi} (dimension inferred from \(\alpha\)) \\
\bottomrule
\end{tabular}
}
\caption{Concrete syntax alternatives for logical constructs. Unicode
alternatives are accepted by both Python syntax and the parser.}
\label{tab:concrete-connectives}
\end{table}

\begin{example}[Formulas in the running example]
\label{ex:video-formulas}
Let \code{sig} be the example signature introduced in
Example~\ref{ex:video-signature}. We also introduce the derived predicate
\(\mathtt{is\_appear}(x)\), defined as \(\neg\,\mathtt{appear}(x,\mathtt{unknown})\), to select genuine appearing videos. Let us consider the following formulas:
\[
\delta :=
\forall (x,y):
\mathtt{appear}(x,y)
\]
\[
\phi :=
\forall (x,y):
\bigl(
\mathtt{appear}(x,y)
\wedge
\mathtt{is\_appear}(x)
\rightarrow
\forall t:
\bigl(
\mathtt{Complete}(x[t])[t]
\rightarrow
\mathtt{digit}(x[t]) =_d y
\bigr)
\bigr)
\]
\[
\psi :=
\forall x:
\bigl(
\mathtt{is\_appear}(x)
\rightarrow
\forall t,t_1 \mid \mathtt{next}(t,t_1):
\bigl(
\mathtt{Complete}(x[t])[t]
\rightarrow
\mathtt{Complete}(x[t_1])[t_1]
\bigr)
\bigr)
\]
\[
\chi :=
\forall x:
\bigl(
\mathtt{is\_appear}(x)
\rightarrow
\forall t:
\bigl(
\neg\,\mathtt{Complete}(x[t])[t]
\rightarrow
\mathtt{digit}(x[t]) =_d \mathtt{unknown}
\bigr)
\bigr)
\]
\[
\beta :=
\forall x:
\bigl(
\mathtt{is\_appear}(x)
\rightarrow
\bigl(
\neg\,\mathtt{Complete}(x[t][t=0])
\wedge
\mathtt{Complete}(x[t][t=l_T-1])
\bigr)
\bigr).
\]
Formula \(\delta\) is the direct supervision clause for the whole-video predicate \(\mathtt{appear}\): notice the use of diagonal  quantifier over video-label pairs \(x,y\). In the running dataset, \(y\) is the drawn digit for genuine appearing-digit videos and \(\mathtt{unknown}\) for non-appearing videos.

Formulas \(\phi\), \(\psi\), \(\chi\) and \(\beta\) are gated by \(\mathtt{is\_appear}\). Formula \(\phi\) relates the whole-video predicate \(\mathtt{appear}\) to frame-level predictions: if \(x\) has label \(y\) and is an appearing video, then every complete frame of \(x\) is assigned digit \(y\). Formula \(\psi\) uses guarded structural quantification over consecutive time indices to express
temporal persistence of \(\mathtt{Complete}\). Formula \(\chi\) assigns the symbol \(\mathtt{unknown}\) to incomplete frames of appearing videos. Formula \(\beta\) uses axis selection to refer to the first and last time steps, stating that an appearing video is incomplete at time \(0\) and complete at time \(l_T-1\).

In \sltn{}, the formulas above can be declared and parsed as follows:
\begin{lstlisting}[language=Python]
sig.define("is_appear", ["x"], "not appear(x, unknown)")
delta = sig.parse("forall (x, y): appear(x, y)")
phi = sig.parse("forall (x, y): ((appear(x, y) & is_appear(x)) ->" 
                  "(forall t: (Complete(x[t])[t] -> (digit(x[t]) =d y))))")
psi = sig.parse("forall x: (is_appear(x) -> "
                  "(forall t,t1 | next(t,t1):(Complete(x[t])[t] -> Complete(x[t1])[t1])))")
chi = sig.parse("forall x: (is_appear(x) -> "
                  "(forall t:(not Complete(x[t])[t] -> (digit(x[t]) =d unknown))))")
beta = sig.parse("forall x: (is_appear(x) -> "
                  "(not Complete(x[t][t=0]) & Complete(x[t][t=l_T-1])))")
\end{lstlisting}

Here, \code{sig.define} is used only to introduce the local shorthand
\(\mathtt{is\_appear}\). In this example, \(\mathtt{appear}(x,\mathtt{unknown})\) is used to identify
non-appearing videos, so \(\mathtt{is\_appear}(x)\) abbreviates its negation. Named definitions will be discussed systematically later. The calls to \code{sig.parse} resolve the
concrete notation to the corresponding \code{Term} or \code{Formula}
object, i.e.\ an abstract syntax tree. The variables \(t\) and \(t_1\) are
structural variables ranging over the temporal dimension. The distinction between \(t\) and \(t_1\) in \(\psi\) is essential: the two indices range over the same structural dimension but must not be aligned.
\end{example}

\paragraph{Operator precedence.}
In the concrete notation, when parentheses are omitted, operator precedence is
as shown in Table~\ref{tab:precedence}. Higher-precedence operators bind more
tightly, and binary connectives of the same precedence associate to the right.
Thus \(\lnot\phi\land\psi\) is read as \((\lnot\phi)\land\psi\),
\(\phi\lor\psi\imp\chi\) as \((\phi\lor\psi)\imp\chi\), and
\(\phi\imp\psi\imp\chi\) as \(\phi\imp(\psi\imp\chi)\).

Quantifiers bind only their immediate subformula. Hence
\(Qx:\phi\circ\psi\) is read as \((Qx:\phi)\circ\psi\), not as
\(Qx:(\phi\circ\psi)\). In particular,
\(\forall x:A(x)\imp B(x)\) means \((\forall x:A(x))\imp B(x)\), whereas
\(\forall x:(A(x)\imp B(x))\) must be written explicitly. Nested quantifiers
associate to the right: \(Qx:Qy:\phi\) is read as \(Qx:(Qy:\phi)\).

\begin{table}[h]
\centering
{\renewcommand{\arraystretch}{1.}
\begin{tabular}{cll}
\toprule
Precedence (tightest first) & Operator(s) & Associativity \\
\midrule
1 & $\lnot$ & -- \\
2 & $\land$ & right \\
3 & $\lor$ & right \\
4 & $\imp$ & right \\
5 & $\leftrightarrow$ & parenthesize / none \\
6 & $\forall,\exists$ & -- \\
\bottomrule
\end{tabular}
\caption{Operator precedence, from tightest to loosest. Binary connectives of
the same precedence are grouped to the right, unless parentheses indicate
otherwise.}
\label{tab:precedence}}
\end{table}

\paragraph{Free variables and free structural variables.} 
Each term and formula has two syntactically defined sets: the set of free
first-order variables, written \(\fv(\cdot)\), and the set of free
structural variables, written \(\fsv(\cdot)\). 

Since variables and constants are declared in the signature together with their dimensional profile, an occurrence of a bare variable or constant may already carry structural axes. If \(a\in\X\cup\C\) and
\[
\dims(a)=\langle d_1,\ldots,d_k\rangle,
\]
then a bare occurrence of \(a\) has the default free structural variables
\[
\dvar{\dims(a)}
=(\alpha_1,\ldots,\alpha_k),
\qquad
\mathit{dim}(\alpha_i)=d_i.
\]
These default variables are generated position-wise from the declared
dimensional profile. If a dimension \(d\) occurs only once, the corresponding
default structural variable is named \(d\). If \(d\) occurs multiple times, its
occurrences are assigned distinct default names \(d_0,d_1,\ldots\), in their
order of occurrence. Thus repeated occurrences of the same dimension are never
collapsed into a single free structural variable.

Function and predicate applications may consume and produce structural axes.
For \(g\in\F\cup\Pr\), the input profile \(\indim(g)\) specifies which
structural axes of the arguments are consumed by the application, while the
output profile \(\outdim(g)\) specifies which structural axes are produced by
the resulting term or formula. Hence
\[
\fsv(g(t_1,\ldots,t_n))
=
\left(
\bigcup_i \fsv(t_i)
\setminus
\mathrm{cons}_g(t_1,\ldots,t_n)
\right)
\cup
\dvar{\outdim(g)}.
\]
Here \(\mathrm{cons}_g(t_1,\ldots,t_n)\) denotes the subset of
\(\bigcup_i\fsv(t_i)\) whose dimensions are matched by the input profile
\(\indim(g)\). If several structural variables have the same dimension, the
matching is performed position-wise according to the ordered profile
\(\indim(g)\). Consumed variables do not remain free in the result unless the
same dimensions are produced again by \(\outdim(g)\), in which case fresh
default variables are introduced through \(\dvar{\outdim(g)}\).

Both \(\fv\) and \(\fsv\) are then defined compositionally on
the abstract syntax tree. For compound expressions, they are obtained by taking
the union of the corresponding sets of the immediate subexpressions and then
removing variables bound by the enclosing first-order or structural binder.
The complete set of rules is given in Table~\ref{tab:freevars}.

\begin{table}[h]
\centering
{\renewcommand{\arraystretch}{1.25}
\resizebox{\linewidth}{!}{%
\begin{tabular}{lll}
\toprule
Expression \(e\) & \(\fv(e)\) & \(\fsv(e)\) \\
\midrule

\(x\), \(\dims(x)=\langle d_1,\ldots,d_k\rangle\) &
\(\{x\}\) &
\(\dvar{\dims(x)}\) \\

\(c\), \(\dims(c)=\langle d_1,\ldots,d_k\rangle\) &
\(\emptyset\) &
\(\dvar{\dims(c)}\) \\

\(g(t_1,\ldots,t_n)\), \(g\in\F\cup\Pr\) &
\(\displaystyle\bigcup_i \fv(t_i)\) &
\(\displaystyle
\left(
\bigcup_i \fsv(t_i)
\setminus
\mathrm{cons}_g(t_1,\ldots,t_n)
\right)
\cup
\dvar{\outdim(g)}\) \\

\(R(\alpha_1,\ldots,\alpha_n)\) &
\(\emptyset\) &
\(\{\alpha_1,\ldots,\alpha_n\}\) \\

\(\lnot\phi\) &
\(\fv(\phi)\) &
\(\fsv(\phi)\) \\

\(\phi\circ\psi\) &
\(\fv(\phi)\cup\fv(\psi)\) &
\(\fsv(\phi)\cup\fsv(\psi)\) \\

\(\lnot\rho\) &
\(\emptyset\) &
\(\fsv(\rho)\) \\

\(\rho\circ\sigma\) &
\(\emptyset\) &
\(\fsv(\rho)\cup\fsv(\sigma)\) \\

\(Q\,x_1,\ldots,x_h : \phi\) &
\(\fv(\phi)\setminus\{x_1,\ldots,x_h\}\) &
\(\fsv(\phi)\) \\

\(Q\,x_1,\ldots,x_h \mid \psi : \phi\) &
\((\fv(\phi)\cup\fv(\psi))\setminus
\{x_1,\ldots,x_h\}\) &
\(\fsv(\phi)\cup\fsv(\psi)\) \\

\(Q\,(x_1,\ldots,x_h) : \phi\) &
\(\fv(\phi)\setminus\{x_1,\ldots,x_h\}\) &
\(\fsv(\phi)\) \\

\(Q\,(x_1,\ldots,x_h) \mid \psi : \phi\) &
\((\fv(\phi)\cup\fv(\psi))\setminus
\{x_1,\ldots,x_h\}\) &
\(\fsv(\phi)\cup\fsv(\psi)\) \\

\(Q\,\alpha_1,\ldots,\alpha_k : \phi\) &
\(\fv(\phi)\) &
\(\fsv(\phi)\setminus\{\alpha_1,\ldots,\alpha_k\}\) \\

\(Q\,\alpha_1,\ldots,\alpha_k \mid \rho : \phi\) &
\(\fv(\phi)\) &
\((\fsv(\phi)\cup\fsv(\rho))\setminus
\{\alpha_1,\ldots,\alpha_k\}\) \\

\(t[\alpha_1,\ldots,\alpha_k]\) &
\(\fv(t)\) &
\(\{\alpha_1,\ldots,\alpha_k\}\) \\

\(\phi[\alpha_1,\ldots,\alpha_k]\) &
\(\fv(\phi)\) &
\(\{\alpha_1,\ldots,\alpha_k\}\) \\

\(t[\alpha=n]\) &
\(\fv(t)\) &
\(\fsv(t)\setminus\{\alpha\}\) \\

\(\phi[\alpha=n]\) &
\(\fv(\phi)\) &
\(\fsv(\phi)\setminus\{\alpha\}\) \\

\bottomrule
\end{tabular}
}}
\caption{Compositional rules for free first-order and structural variables.
Here \(\circ\in\{\land,\lor,\imp,\leftrightarrow\}\), \(Q\in\{\forall,\exists\}\),
\(g\in\F\cup\Pr\), \(t\) ranges over terms, \(\phi,\psi\) over formulas, and
\(\rho,\sigma\) over structural formulas. For a dimensional profile
\(\langle d_1,\ldots,d_k\rangle\), the tuple
\(\dvar{\langle d_1,\ldots,d_k\rangle}\) denotes the default structural
variables generated position-wise from the profile.}
\label{tab:freevars}
\end{table}

\begin{example}[Free first-order and structural variables]
\label{ex:free-vars}
Consider the structural subformula
\[
\theta :=
\forall t,t_1 \mid \mathrm{next}(t,t_1) :
\bigl(\mathrm{Complete}(x[t])[t] \rightarrow
\mathrm{Complete}(x[t_1])[t_1]\bigr).
\]
At the atomic level, we have
\[
\fv(\mathrm{Complete}(x[t])[t])=\{x\},
\qquad
\fsv(\mathrm{Complete}(x[t])[t])=\{t\},
\]
and
\[
\fv(\mathrm{Complete}(x[t_1])[t_1])=\{x\},
\qquad
\fsv(\mathrm{Complete}(x[t_1])[t_1])=\{t_1\}.
\]
Moreover, the structural guard satisfies
\[
\fv(\mathrm{next}(t,t_1))=\emptyset,
\qquad
\fsv(\mathrm{next}(t,t_1))=\{t,t_1\}.
\]
Therefore, the inner implication has free variable set \(\{x\}\) and free structural variables set \(\{t,t_1\}\).
The guarded structural quantifier binds \(t\) and \(t_1\), including their
occurrences in the guard. Hence the whole formula satisfies
\[
\fv(\theta)=\{x\},
\qquad
\fsv(\theta)=\emptyset.
\]
\end{example}

In \sltn{}, we can easily access the sets \(\fv(e)\) and \(\fsv(e)\) of a term or formula object \(e\) as follows:
\begin{lstlisting}[language=Python]
e.free_variables
e.free_structural_variables
\end{lstlisting}

\paragraph{Axis selection.}
A term or formula of the form \(e[\alpha=n]\) selects the fixed index
\(n\in\mathbb{N}\) along the structural axis named by \(\alpha\).
Syntactically, this removes \(\alpha\) from the free structural variables of
\(e\). In the running example, axis selection appears in the boundary-condition
clause \(\beta\), within the subformula
\[
\neg\,\mathrm{Complete}(x[t][t=0])
\ \wedge\
\mathrm{Complete}(x[t][t=l_T-1]).
\]
which fixes the temporal axis to the first and last frames of the video. The
clause therefore states that every appearing video starts incomplete and ends
complete. In particular,
\[
\fv(\mathrm{Complete}(x[t][t=0]))=\{x\},
\qquad
\fsv(\mathrm{Complete}(x[t][t=0]))=\emptyset.
\]

\paragraph{Knowledge bases.}
A formula \(\theta\) is a \emph{clause} if both \(\fv(\theta)\) and
\(\fsv(\theta)\) are empty. A \emph{knowledge base} is a finite collection of
clauses. Intuitively, a knowledge base represents the facts, constraints, or
domain knowledge that should be jointly satisfied. Knowledge bases should not
be read merely as abbreviations for single conjunctions: in the semantics and
training interface, clauses may be kept as distinct objectives and combined by
configurable aggregation or multi-objective mechanisms, as discussed in
Section~\ref{s:semantics}.

In concrete syntax, given already constructed clause formula objects
\(\delta,\phi,\psi,\chi,\beta\) as in Example~\ref{ex:video-formulas},
one may construct a knowledge base as follows:
\begin{lstlisting}[language=Python]
kb = sltn.KB(delta,phi,psi,chi,beta)
\end{lstlisting}
The \sltn{} parser also allows brace notation \code{\{\dots\}} as a direct
way to write knowledge bases from formula strings. For instance,
\begin{lstlisting}[language=Python]
kb = sig.parse("{"
    "forall(x,y): appear(x, y)," # delta
    "forall x:(is_appear(x)->(not Complete(x[t][t=0]) & Complete(x[t][t=l_T-1])))" # beta
    "}"
)
\end{lstlisting}
is equivalent to
\begin{lstlisting}[language=Python]
kb = sltn.KB(delta, beta)
\end{lstlisting}

\paragraph{Definitions.}
In addition to primitive function and predicate symbols declared in the
signature, \sltn{} supports \emph{definitions}, i.e., derived terms or formulas
introduced as syntactic abbreviations. A definition associates a fresh symbol
name with a parametrized term or formula body, optionally together with sort
and structural-dimension annotations for its formal parameters. Each occurrence
of the defined symbol is expanded by the parser, substituting the actual
arguments for the formal parameters before semantic interpretation. Definitions
provide a convenient mechanism for naming recurring patterns, derived
predicates, and reusable clauses.

For instance, the auxiliary predicate \(\mathtt{is\_appear}\) in
Example~\ref{ex:video-formulas} is introduced in \sltn{} as follows:
\begin{lstlisting}[language=Python]
sig.define("is_appear", arg_names=["x"],
           formula_body="not appear(x, unknown)")
\end{lstlisting}
\subsection{Semantics}
\label{s:semantics}

The syntax introduced above specifies the well-formed expressions of \sltn{}.
Their meaning is defined relative to an interpretation \(\I\) over a signature
\[
\Sig=(\S,\D,\C,\X,\F,\Pr,\SV,\SR,\sort,\dims).
\]
An interpretation assigns semantic objects to the symbols declared in the
signature: tensor types to sorts, tensors to constants and variables, tensor
maps to functions and predicates, fuzzy masks to structural relations, and fuzzy truth functions and aggregation operators to the logical symbols.

These assignments determine, by structural recursion, the denotation of every
term, formula, and structural formula. Denotations are represented as
\emph{annotated tensors}, namely tensors whose axes are explicitly named and
assigned semantic roles.

We use square brackets for symbol groundings and parentheses for expression
denotations. Thus, \(\I[a]\) denotes the grounding of a signature symbol \(a\),
whereas \(\I(e)\) denotes the denotation of an expression \(e\). For example,
\(\I[x]\) is the raw tensor assigned to the variable symbol \(x\), whereas
\(\I(x)\) is the annotated tensor denoted by the variable term \(x\).

In \sltn{}, once a signature \code{sig} has been declared, an interpretation \code{interp} over \code{sig} is instantiated as follows:
\begin{lstlisting}[language=Python]
interp = Interpretation(sig)
\end{lstlisting}

\subsubsection{Signature Grounding}
\label{s:signature-grounding}

The first semantic level assigns raw objects to the symbols declared in the
signature: types to sorts, tensors to constants and variables, tensor maps to
functions and predicates, and masks to structural relations. We call this operation \emph{grounding}.

\paragraph{Sorts.}
Each sort \(s\in\S\) is grounded to a \emph{type}
\(\tau(s)\), that is, a subset
\[
\I[s]=\tau(s)
\subseteq
\mathbb{R}^{n_1\times\cdots\times n_d},
\qquad n_i\in\mathbb{N}.
\]
We write
\[
|\tau|=(n_1,\ldots,n_d),
\qquad
\mathbb{R}^{|\tau|}
=
\mathbb{R}^{n_1\times\cdots\times n_d}.
\]

\noindent In \sltn{}, a type is specified by a tuple
\[
\tau \sim
(\mathit{name},\mathit{shape},\mathit{axis\ names},\mathit{constraint}),
\]
where \(\mathit{name}\) is a symbolic name, \(\mathit{shape}\) is the tensor
shape \(|\tau|\), \(\mathit{axis\ names}\) optionally names the corresponding
domain axes, and \(\mathit{constraint}\) optionally restricts the admissible
tensors. Such a tuple determines the domain
\[
\tau
=
\left\{
\mathbf{x}\in\mathbb{R}^{|\tau|}
\bnfalt
\mathit{constraint}(\mathbf{x})
\right\}.
\]
If no constraint is specified, then \(\tau=\mathbb{R}^{|\tau|}\).

The built-in sorts \code{Real} and \code{Bool} are grounded by default
types. The sort \code{Real} is represented by scalar real-valued tensors,
whereas \code{Bool} is represented by fuzzy truth degrees in \([0,1]\):
\begin{lstlisting}[language=Python]
Type(name="Real", shape=(1,), axis_names=("real",))
Type(name="Bool", shape=(1,), axis_names=("bool",),
     constraint=lambda x: (x >= 0).all() and (x <= 1).all())
\end{lstlisting}

In the running example, frames are represented by tensors of shape
\(1\times 28\times 28\), with axes corresponding to channel, height, and width.
Digits are represented by vectors of length \(11\), including the
\(\mathtt{unknown}\) class. The interpretation grounds the signature sorts as
follows:
\begin{lstlisting}[language=Python]
interp["Image"] = Type("Image", shape=(1, 28, 28),
                         axis_names=("channel", "height", "width"))
interp["Digit"] = Type("Digit", shape=(11,), axis_names=("digit",))
\end{lstlisting}

\paragraph{Constants and variables.}
Let \(c\in\C\) be a constant of sort \(s=\sort(c)\), and let
\[
\dims(c)=\langle d_1,\ldots,d_r\rangle
\]
be its declared dimensional profile, possibly with repetitions. A constant
\(c\) is grounded to a tensor
\[
\I[c]
\in
\mathbb{R}^{\ell_{d_1}\times\cdots\times \ell_{d_r}\times |\tau(s)|},
\]
where \(\tau(s)=\I[s]\), each \(\ell_{d_i}\in\mathbb{N}\) is the finite extent of the corresponding structural dimension in the current evaluation context, and the final axes are the domain axes of \(\tau(s)\). If \(\dims(c)=\langle\rangle\), then
\[\I[c]\in\mathbb{R}^{|\tau(s)|}.\]

Variables are grounded analogously, with an additional leading variable axis.
Let \(x\in\X\) be a variable of sort \(s=\sort(x)\), and let
\[
\dims(x)=\langle d_1,\ldots,d_r\rangle.
\]
Then \(x\) is grounded to a tensor
\[
\I[x]
\in
\mathbb{R}^{b_x\times
\ell_{d_1}\times\cdots\times \ell_{d_r}\times |\tau(s)|}.
\]
The leading axis has size \(b_x\) and ranges over the individuals currently
assigned to the first-order variable \(x\), for instance the elements of a
minibatch. The following \(r\) axes correspond to the declared structural
dimensions, and the final axes are the domain axes of \(\tau(s)\).

At the raw grounding level, structural axes are ordered according to the
declared dimensional profile. In expression denotations, these axes are named
by structural variables, either by default variables generated from the profile
or by explicit annotations such as \(x[t]\).

In the running example, \(x\) has sort \code{Image} and
\(\dims(x)=\langle\mathtt{T}\rangle\). For a minibatch \(B_x\) of \(32\)
videos of length \(8\), one may ground \(x\) by
\[
\I[x]=B_x
\in
\mathbb{R}^{32\times 8\times 1\times 28\times 28}.
\]
The first axis ranges over the \(32\) video individuals assigned to \(x\), the
second axis corresponds to the temporal dimension \(\mathtt{T}\), and the
remaining axes are the image-domain axes. The variable \(y\) has sort
\code{Digit} and no structural dimension; hence, for a minibatch \(B_y\) of
labels,
\[
\I[y]=B_y\in\mathbb{R}^{32\times 11}.
\]
The constant \(\mathtt{unknown}\) has sort \code{Digit} and no structural
dimension:
\[
\I[\mathtt{unknown}]\in\mathbb{R}^{11}.
\]
In the implementation:
\begin{lstlisting}[language=Python]
interp["x"] = B_x
interp["y"] = B_y
interp["unknown"] = torch.tensor([0., 0., ..., 1.])  # one-hot vector
\end{lstlisting}

\paragraph{Functions.}
Let \(f\in\F\) have input and output sorts
\[
\sort_{\mathrm{in}}(f)=\langle s_1,\ldots,s_m\rangle,
\qquad
\outsort(f)=s_0,
\]
and input and output dimensional profiles
\[
\indim(f)=\langle d_1,\ldots,d_r\rangle,
\qquad
\outdim(f)=\langle e_1,\ldots,e_q\rangle.
\]
Let \(\I[s_j]=\tau_j\), for \(j=0,\ldots,m\). In an evaluation
context where the corresponding structural dimensions have extents
\[
\ell_d=(\ell_{d_1},\ldots,\ell_{d_r}),
\qquad
\ell_e=(\ell_{e_1},\ldots,\ell_{e_q}),
\]
a grounding for \(f\) is a tensor map
\[
\I[f]:
\left(
\mathbb{R}^{|\tau_1|}
\times\cdots\times
\mathbb{R}^{|\tau_m|}
\right)^{\ell_d}
\longrightarrow
\left(
\mathbb{R}^{|\tau_0|}
\right)^{\ell_e}.
\]
Here \(V^{\ell_d}\) abbreviates an array of \(V\)-valued elements indexed by
the product of the extents in \(\ell_d\). If both structural profiles are empty, this reduces to a local map
\[
\I[f]:
\mathbb{R}^{|\tau_1|}
\times\cdots\times
\mathbb{R}^{|\tau_m|}
\longrightarrow
\mathbb{R}^{|\tau_0|}.
\]
\noindent In practice, the type above describes the common representation passed to the grounding after alignment and broadcasting, as we will see in detail in the next section. It does not require every argument to carry all dimensions in \(\indim(f)\) natively. 


In the running example, the function \(\mathtt{digit}\) has empty input and
output structural profiles:
\[
\indim(\mathtt{digit})=\outdim(\mathtt{digit})=\langle\rangle.
\]
It is therefore grounded by a local frame-level classifier
\[
\I[\mathtt{digit}]:
\mathbb{R}^{1\times 28\times 28}
\longrightarrow
\mathbb{R}^{11}.
\]
When applied to the structurally indexed term \(x[t]\), the temporal axis is
external to the local grounding and is propagated pointwise.
\begin{lstlisting}[language=Python]
interp["digit"] = DigitClassifier() # torch.nn.Module()
\end{lstlisting}

\paragraph{Predicates.}
Predicates are grounded analogously to functions, except that their output type
is the Boolean type. Let \(p\in\Pr\) have input sorts
\[
\sort_{\mathrm{in}}(p)=\langle s_1,\ldots,s_m\rangle
\]
and dimensional profiles
\[
\indim(p)=\langle d_1,\ldots,d_r\rangle,
\qquad
\outdim(p)=\langle e_1,\ldots,e_q\rangle.
\]
If \(\I[s_j]=\tau_j\) and the relevant dimensional extents are \(\ell_d\) and \(\ell_e\), then a grounding of \(p\) is a truth-valued tensor map
\[
\I[p]:
\left(
\mathbb{R}^{|\tau_1|}
\times\cdots\times
\mathbb{R}^{|\tau_m|}
\right)^{\ell_d}
\longrightarrow
[0,1]^{\ell_e}.
\]
Thus, a predicate returns fuzzy truth degrees, possibly indexed by its declared
output structural dimensions.

In the running example, \(\mathtt{Complete}\) has empty input and output
structural profiles. It is grounded by a local frame-level truth-valued map
\[
\I[\mathtt{Complete}]:
\mathbb{R}^{1\times 28\times 28}
\longrightarrow
[0,1].
\]
When applied to \(x[t]\), the temporal axis is propagated pointwise by the
compositional semantics. The equality predicate \(=_d\) is grounded by a fuzzy similarity map on digit representations:
\[
\I[=_d]:
\mathbb{R}^{11}\times\mathbb{R}^{11}
\longrightarrow
[0,1].
\]
The predicate \(\mathtt{appear}\), in contrast, consumes the temporal
dimension:
\[
\indim(\mathtt{appear})=\langle\mathtt{T}\rangle,
\qquad
\outdim(\mathtt{appear})=\langle\rangle.
\]
Its grounding has type
\[
\I[\mathtt{appear}]:
\left(
\mathbb{R}^{1\times 28\times 28}
\times
\mathbb{R}^{11}
\right)^{\ell_{\mathtt{T}}}
\longrightarrow
[0,1].
\]
Operationally, this may be implemented by a sequence-level classifier that
receives the whole video \(x\), aligned along the consumed temporal axis,
together with a candidate label \(y\). Since \(y\) has no temporal axis, it is
broadcast along \(\mathtt{T}\) before the predicate grounding is applied.

In \sltn{}, the groundings may be specified as follows:
\begin{lstlisting}[language=Python]
interp["Complete"] = CompleteClassifier() # torch.nn.Module
interp["appear"] = AppearClassifier() # recurrent model
interp["=d"] = lambda a, b: (a * b).sum(dim=-1, keepdim=True) # Python callable
\end{lstlisting}

\paragraph{Structural relations.}
Structural relations are grounded by masks over structural positions. Let
\(R\in\SR\) be a structural relation with dimensional profile
\[
\dims_{\mathrm{rel}}(R)=\langle d_1,\ldots,d_k\rangle.
\]
If, in the current evaluation context, the corresponding dimensions have
finite extents
\[
\ell_d=(\ell_{d_1},\ldots,\ell_{d_k}),
\]
then the grounding of \(R\) is a truth-valued mask
\[
\I[R]\in
[0,1]^{\ell_{d_1}\times\cdots\times\ell_{d_k}}.
\]
The entry \(\I[R]_{i_1,\ldots,i_k}\) specifies the degree, or weight,
to which the tuple of structural positions \((i_1,\ldots,i_k)\) satisfies the
relation. A structural relation is said to be \emph{crisp} if all mask values
belong to \(\{0,1\}\). 

For the temporal relation \(\mathtt{next}\) in the running example,
\[
\dims_{\mathrm{rel}}(\mathtt{next})
=
\langle\mathtt{T},\mathtt{T}\rangle.
\]
If the temporal extent is \(8\), then a crisp interpretation is given by
\[
\I[\mathtt{next}]_{ij}
=
\delta_{i,j-1}
\in
\{0,1\}^{8\times 8},
\]
where \(\delta\) is the Kronecker delta. Thus
\(\I[\mathtt{next}]_{ij}=1\) precisely when \(j=i+1\). In \sltn{},
the mask can be instantiated as follows:
\begin{lstlisting}[language=Python]
interp["next"] = torch.diag(torch.ones(8 - 1), diagonal=1)
\end{lstlisting}

\begin{figure}[h]
\centering
\includegraphics[width=.6\linewidth]{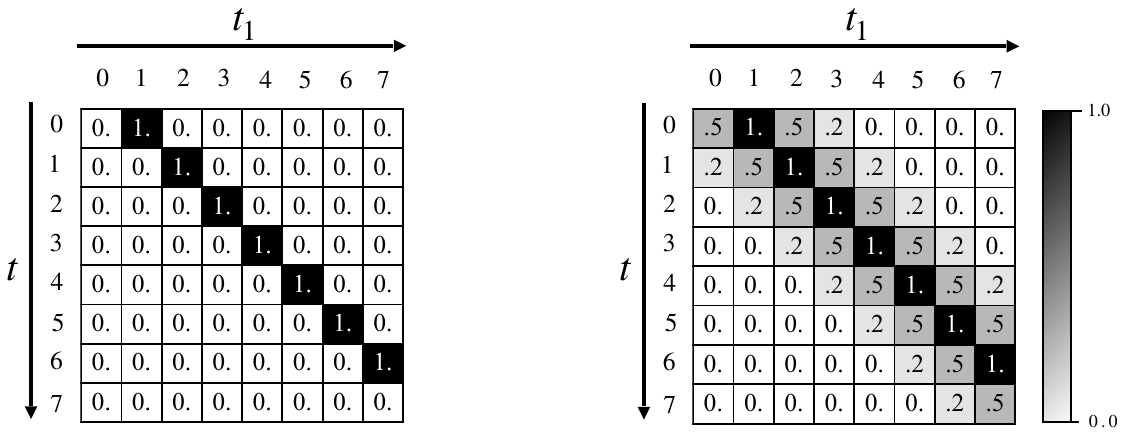}
\caption{Crisp, left, and soft, right, interpretations of the
\(\mathtt{next}(t,t_1)\) structural relation.}
\label{fig:structural-masks}
\end{figure}

\paragraph{Logical operators.}
Logical connectives and quantifiers are grounded by fuzzy operators and aggregators. In general, negation is interpreted by a fuzzy negator \(N\), conjunction by a t-norm \(T\), disjunction by a t-conorm \(S\), implication by a fuzzy implication \(J\), and equivalence by a fuzzy equivalence operator
\(E\):
\[
N:[0,1]\to[0,1],
\qquad
T,S,J,E:[0,1]^2\to[0,1].
\]
Standard configurations may instantiate these operators using established families of fuzzy logics, such as Gödel, Goguen, or \L{}ukasiewicz logic. Differentiable operators that relax operator properties, such as associativity or the existence of a neutral element, are also admissible, provided that they remain defined on \([0,1]\) and coincide with the corresponding Boolean connectives on crisp truth values. Definitions, algebraic properties, and default choices are given in Appendix~\ref{a:operators}.

Universal and existential quantifiers are interpreted by aggregation operators
\[
A_Q:
\bigcup_{n\in\mathbb{N}} [0,1]^n \to [0,1],
\qquad
Q\in\{\forall,\exists\}.
\]
Structural quantifiers are likewise grounded by aggregation operators, but then applied over structural axes. Each structural dimension may be assigned an independent grounding. Guarded quantifiers use the same aggregators together with a mask or weight tensor supplied by the guard. More precisely, a guarded quantifier is grounded by a masked aggregator
\[
A_Q^{\mathrm{mask}}:
\bigcup_{n\in\mathbb{N}}
\Big([0,1]\times\underbrace{[0,1]}_{\text{weight}}\Big)^n
\to
[0,1],
\qquad
Q\in\{\forall,\exists\}.
\]
The way in which the guard is applied to the data is described in Appendix~\ref{a:operators}. All aggregators can be applied to crisp masks by restricting the aggregation to the selected elements; soft guarded quantification, however, is supported only by selected aggregators.

Different operators and aggregators induce different gradient-flow behaviour and should therefore be chosen according to the task. A possible configuration in \sltn{} is:
\begin{lstlisting}[language=Python]
interp.logic = Logic()
interp.logic["and"] = ops.AndLuk()  # max(x + y - 1, 0)
interp.logic["not"] = ops.NotStandard()  # Lukasiewicz: 1 - x
interp.logic = interp.logic.with_defaults()  # completes the assignment to all operators
interp.logic["forall"] = ops.aggregator.AggregPMean(p=1/5)
interp.logic["forall,T"] = ops.aggregator.AggregPMeanError(p=10)
\end{lstlisting}

The last two assignments provide distinct groundings for universal quantification over variable axes and over the structural dimension \code{T}, respectively.

\vspace{10pt}
At this point, all symbols declared in the signature have been assigned raw semantic objects. Table~\ref{tab:interpretation-api} summarizes the signature groundings for the running example. The next section explains how these groundings define the denotations of terms and formulas as annotated tensors.

\begin{table}[h]
\centering
\renewcommand{\arraystretch}{1.08}
\resizebox{\linewidth}{!}{%
\begin{tabular}{lll}
\toprule
\textbf{Grounded component} &
\textbf{Semantic role} &
\textbf{\sltn{} example code} \\
\midrule

sort \(s\in\S\) &
type / admissible tensor domain &
\code{interp["Image"] = ImageType} \\
& &
\code{interp["Digit"] = DigitType} \\

constant \(c\in\C\) &
tensor with structural and domain axes &
\code{interp["unknown"] = unknown\_vec} \\

variable \(x\in\X\) &
tensor with variable, structural, and domain axes &
\code{interp["x"] = B\_x} \\
& &
\code{interp["y"] = B\_y} \\

function \(f\in\F\) &
local tensor map &
\code{interp["digit"] = DigitClassifier()} \\

predicate \(p\in\Pr\) &
truth-valued tensor map &
\code{interp["Complete"] = CompleteClassifier()} \\
& &
\code{interp["appear"] = AppearClassifier()} \\
& &
\code{interp["=d"] = digit\_similarity} \\

structural relation \(R\in\SR\) &
mask over structural indices &
\code{interp["next"] = next\_mask} \\

logical operators &
fuzzy operators and aggregators &
\code{interp.logic = Logic()} \\
& &
\code{interp.logic["forall,T"] = ...} \\

\bottomrule
\end{tabular}%
}
\caption{Summary of signature groundings for the running example.}
\label{tab:interpretation-api}
\end{table}

\subsubsection{Denotation of Terms and Formulas}
\label{s:compositional-semantics}

The raw groundings introduced above are lifted to denotations of compound terms, formulas, and structural formulas by structural recursion. These denotations are represented as \emph{annotated tensors}, which make explicit the semantic role of each tensor axis and thereby support compositional operations such as named-axis alignment, broadcasting, local map application, axis selection, renaming, and aggregation.

\paragraph{Annotated tensors.}
An \emph{annotated tensor} is a tuple
\[
e=(\val{e},\axes(e),\tau(e)),
\]
where \(\val{e}\) is a tensor, \(\axes(e)\) is an ordered list assigning a name and a
role to each axis of \(v\), and \(\tau(e)\) is the type of the expression. The
possible axis roles are \emph{variable}, \emph{structural}, and
\emph{domain}.

Every well-formed term or formula is denoted by an annotated tensor whose
variable axes are determined by its free first-order variables, whose
structural axes are determined by its free structural variables, and whose
domain axes are determined by its type. More precisely, let \(q\) be a term of
sort \(s\). Let \(x_1,\ldots,x_m\) be the distinct free first-order variables
of \(q\), and let \(\alpha_1,\ldots,\alpha_r\) be the distinct free structural
variables of \(q\). If \(\I[s]=\tau(s)\), and if
\(a_1,\ldots,a_k\) are the domain-axis names of \(\tau(s)\), then
\[
\I(q)
=
(\val{q},\axes(q),\tau(s)),
\]
where
\[
\val{q}
\in
\mathbb{R}^{
b_{x_1}\times\cdots\times b_{x_m}
\times
\ell_{\alpha_1}\times\cdots\times\ell_{\alpha_r}
\times
|\tau(s)|
},
\]
and, up to the chosen canonical ordering of free axes,
\[
\axes(q)
=
\bigl[
(x_1,\mathrm{var}),\ldots,(x_m,\mathrm{var}),
(\alpha_1,\mathrm{str}),\ldots,(\alpha_r,\mathrm{str}),
(a_1,\mathrm{dom}),\ldots,(a_k,\mathrm{dom})
\bigr].
\]

For a formula \(\phi\), the same convention applies, except that its type is
\(\tau(\code{Bool})\). Thus,
\[
\I(\phi)
=
(\val{\phi},\axes(\phi),\tau(\code{Bool})),
\]
where \(\val{\phi}\) carries the variable axes associated with
\(\fv(\phi)\), the structural axes associated with
\(\fsv(\phi)\), and the Boolean domain axis. Its entries are fuzzy
truth degrees in \([0,1]\).

For example, let \(x\) be a variable of sort \code{Image} indexed by a
temporal structural dimension, and suppose that \(x\) is grounded by a batch
\(B_x\) of \(32\) videos of length \(8\). Evaluating \(\I(x)\) in
\sltn{} yields an annotated tensor of the form:
\begin{lstlisting}
Tensor(shape=(x(variable): 32, T(structural): 8,
        channel(domain): 1, height(domain): 28, width(domain): 28),
        domain_type=Image)
\end{lstlisting}
After an explicit annotation such as \(x[t]\), the temporal structural axis is named by the structural variable \(t\). We now define these denotations by structural recursion, following the grammar
of terms and formulas.

\paragraph{Named-axis alignment and broadcasting.}
Most compositional rules require several annotated tensors to be brought to a common free-axis shape before a pointwise operator or a local tensor map is
applied. This is done by named-axis alignment. Given annotated tensors
\(e_1,\ldots,e_n\), we write
\[
\operatorname{align}(e_1,\ldots,e_n)
=
(\widehat e_1,\ldots,\widehat e_n)
\]
for the operation that constructs a common ordered list \(\Gamma\) containing all variable and structural axes occurring in the inputs, identified by name and role. Each tensor is reshaped by inserting singleton axes for the axes in \(\Gamma\) that it does not carry, and its underlying tensor is broadcast to the corresponding common free-axis shape. Domain axes are not aligned.

\paragraph{Denotation of terms.}
Terms are interpreted by structural recursion over the term grammar.
\begin{enumerate}[label=(\roman*), itemsep=4pt]
    \item \emph{Constants.}
    Let \(c\) be a constant of sort \(s=\sort(c)\), and let
    \(\dims(c)=\langle d_1,\ldots,d_r\rangle.\)
    Let
    \(\dvar{\dims(c)}=(\alpha_1,\ldots,\alpha_r)\)
    be the default structural variables generated from the declared
    dimensional profile \(\dims(c)\), with
    \[
    \mathit{dim}(\alpha_i)=d_i
    \qquad
    \text{for } i=1,\ldots,r.
    \]
    If \(\I[s]=\tau(s)\), and if \(a_1,\ldots,a_k\) are the
    domain-axis names of \(\tau(s)\), then the raw grounding \(\I[c]\)
    is turned into the annotated tensor
    \[
    \I(c)
    =
    \bigl(\I[c],\axes(c),\tau(s)\bigr).
    \]
    Its axes consist of the default structural variables generated from
    \(\dims(c)\), followed by the domain axes of \(\tau(s)\). Thus, if
    \[
    \I[c]\in
    \mathbb{R}^{\ell_{d_1}\times\cdots\times \ell_{d_r}\times |\tau(s)|},
    \]
    then
    \[
    \axes(c)
    =
    \bigl[
    (\alpha_1,\mathrm{str}),\ldots,(\alpha_r,\mathrm{str}),
    (a_1,\mathrm{dom}),\ldots,(a_k,\mathrm{dom})
    \bigr].
    \]
    If \(\dims(c)=\langle\rangle\), then no structural axes are present and
    \[
    \axes(c)
    =
    \bigl[
    (a_1,\mathrm{dom}),\ldots,(a_k,\mathrm{dom})
    \bigr].
    \]
    \item \emph{Variables.}
    Variables are interpreted similarly, with an additional leading variable
    axis. Let \(x\) be a variable of sort \(s=\sort(x)\), and let
    \(\dims(x)=\langle d_1,\ldots,d_r\rangle.\)
    Let
    \(\dvar{\dims(x)} = (\alpha_1,\ldots,\alpha_r)\)
    be the default structural variables generated from the declared
    dimensional profile \(\dims(x)\), with
    \[
    \mathit{dim}(\alpha_i)=d_i
    \qquad
    \text{for } i=1,\ldots,r.
    \]
    If \(x\) is grounded by
    \[
    \I[x]\in
    \mathbb{R}^{b_x\times
    \ell_{d_1}\times\cdots\times \ell_{d_r}\times |\tau(s)|},
    \]
    then
    \[
    \I(x)
    =
    \bigl(\I[x],\axes(x),\tau(s)\bigr),
    \]
    with
    \[
    \axes(x)
    =
    \bigl[
    (x,\mathrm{var}),
    (\alpha_1,\mathrm{str}),\ldots,(\alpha_r,\mathrm{str}),
    (a_1,\mathrm{dom}),\ldots,(a_k,\mathrm{dom})
    \bigr].
    \]
    The first axis ranges over the individuals assigned to the variable \(x\);
    the following axes are named by the default structural variables generated
    from \(\dims(x)\); and the final axes are the domain axes of its sort.
    If \(\dims(x)=\langle\rangle\), then no structural axes are present and
    \[
    \axes(x)
    =
    \bigl[
    (x,\mathrm{var}),
    (a_1,\mathrm{dom}),\ldots,(a_k,\mathrm{dom})
    \bigr].
    \] 
    \item \emph{Function applications.}
    Consider a function application
    \[
    f(q_1,\ldots,q_m).
    \]
    Assume that each argument has already been interpreted as an annotated
    tensor
    \[
    \I(q_j)=(v_j,\axes(q_j),\tau_j),
    \qquad j=1,\ldots,m.
    \]
    Let
    \[
    \sort_{\mathrm{in}}(f)=\langle s_1,\ldots,s_m\rangle,
    \qquad
    \outsort(f)=s_0,
    \]
    and let
    \[
    \indim(f)=\Delta_{\mathrm{in}}
    =
    \langle d_1,\ldots,d_r\rangle,
    \qquad
    \outdim(f)=\Delta_{\mathrm{out}}
    =
    \langle e_1,\ldots,e_q\rangle.
    \]
    Finally, let \(\tau_0=\I[s_0]\), and let
    \(a_1,\ldots,a_k\) be the domain-axis names of \(\tau_0\).

    The input structural profile \(\Delta_{\mathrm{in}}\) specifies the
    structural dimensions consumed by the application of \(f\). These
    dimensions are consumed jointly from the arguments after named-axis
    alignment and broadcasting. More precisely, the argument denotations are
    first aligned by name and role on their free variable and structural axes:
    \[
    \operatorname{align}
    \bigl(
    \I(q_1),\ldots,\I(q_m)
    \bigr)
    =
    (\widehat q_1,\ldots,\widehat q_m).
    \]
    This alignment identifies axes with the same name and role, inserts
    singleton axes where an argument does not carry a free axis present in
    another argument, and broadcasts the corresponding tensor values to the
    common free-axis shape.

    Among the aligned structural axes, those whose structural dimensions occur
    in \(\Delta_{\mathrm{in}}\) are treated as the consumed structural axes of
    the application. If an argument already carries such an axis, it contributes a genuinely indexed family of values along that axis. If an argument does not carry such an axis, it is broadcast along it.

    The grounding
    \[
    \I[f]:
    \left(
    \mathbb{R}^{|\tau_1|}
    \times\cdots\times
    \mathbb{R}^{|\tau_m|}
    \right)^{\ell_{d_1}\times\cdots\times\ell_{d_r}}
    \longrightarrow
    \left(
    \mathbb{R}^{|\tau_0|}
    \right)^{\ell_{e_1}\times\cdots\times\ell_{e_q}}
    \]
    is then applied to the aligned consumed structural axes and to the domain
    axes of the arguments. Variable axes and structural axes not consumed by
    \(\Delta_{\mathrm{in}}\) are external axes; they are propagated pointwise
    through the application.

    Let \(\dvar{\outdim(f)} = (\beta_1,\ldots,\beta_q)\) be the default structural variables generated for the output structural profile \(\outdim(f)\), with
    \[
    \mathit{dim}(\beta_i)=e_i
    \qquad
    \text{for } i=1,\ldots,q.
    \]
    The result is an annotated tensor
    \[
    \I\bigl(f(q_1,\ldots,q_m)\bigr)
    =
    \bigl(
    v_f,\Gamma_f,\tau_0
    \bigr),
    \]
    where \(v_f\) is the tensor value returned by the grounding \(\I[f]\), and
    \(\Gamma_f\) is obtained by concatenating, in the chosen canonical order, propagated external variable axes, propagated external structural axes, output structural axes, output domain axes.
    Thus,
    \[
    \Gamma_f
    =
    \Gamma_{\mathrm{var}}
    \mathbin{\Vert}
    \Gamma_{\mathrm{str}}^{\mathrm{ext}}
    \mathbin{\Vert}
    \bigl[
    (\beta_1,\mathrm{str}),\ldots,(\beta_q,\mathrm{str})
    \bigr]
    \mathbin{\Vert}
    \bigl[
    (a_1,\mathrm{dom}),\ldots,(a_k,\mathrm{dom})
    \bigr],
    \]
    where \(\Gamma_{\mathrm{var}}\) contains the propagated variable axes and
    \(\Gamma_{\mathrm{str}}^{\mathrm{ext}}\) contains the propagated structural
    axes not consumed by \(\indim(f)\).

    If \(\outdim(f)=\langle\rangle\), then no output structural axes are added.
    If \(\indim(f)=\langle\rangle\), then no structural axes are consumed, and
    all structural axes occurring freely in the arguments are propagated
    pointwise.
\end{enumerate}

\begin{figure}[h]
\centering
\includegraphics[width=\linewidth]{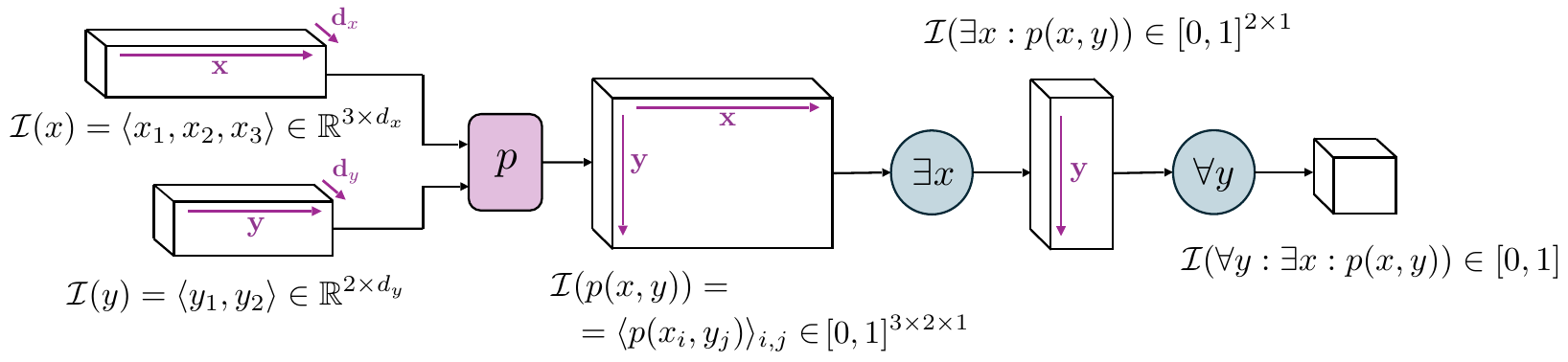}
\caption{Interpretation of the atomic formula
\(p(x,f(x,y))\), adapted from~\citep{badreddine_logic_2022}.}
\label{fig:function-application}
\end{figure}

\paragraph{Axis annotation and axis selection.}
Let \(e\) be a term or formula whose denotation carries output structural axes
\(\beta_1,\ldots,\beta_k\), in the order determined by the expression. The
annotation
\[
e[\alpha_1,\ldots,\alpha_k]
\]
is well formed when
\[
\mathit{dim}(\alpha_i)=\mathit{dim}(\beta_i)
\qquad
\text{for all }i=1,\ldots,k,
\]
and when the variables \(\alpha_1,\ldots,\alpha_k\) are pairwise distinct and
fresh with respect to the remaining free structural variables of \(e\). Its
denotation is obtained from \(\I(e)\) by replacing, in the axis
annotation, each structural-axis name \(\beta_i\) with \(\alpha_i\). The order,
extent, and role of the renamed axes are preserved, as are the expression type
and the underlying tensor values. 

Similarly, if \(\I(e)\) carries a structural axis named \(\alpha\) of
extent \(\ell_\alpha\), then, for \(0\le n<\ell_\alpha\), the expression
\[
e[\alpha=n]
\]
is interpreted by selecting the subtensor obtained by fixing the structural
axis \(\alpha\) to index \(n\). The selected axis is removed from the axis
annotation and from the underlying tensor shape, while the relative order,
names, and roles of all remaining axes, as well as the expression type, are
preserved.

\paragraph{Denotation of structural formulas.}
Structural formulas are interpreted as truth-valued annotated tensors whose
free axes are structural axes. Let
\[
\rho = R(\alpha_1,\ldots,\alpha_k)
\]
be an atomic structural formula, with
\[
\dims_{\mathrm{rel}}(R)=\langle d_1,\ldots,d_k\rangle,
\qquad
\mathit{dim}(\alpha_i)=d_i
\quad
\text{for } i=1,\ldots,k.
\]
Then the raw mask \(\I[R]\) is turned into an annotated tensor by naming its
structural axes with \(\alpha_1,\ldots,\alpha_k\):
\[
\I(R(\alpha_1,\ldots,\alpha_k))
=
\bigl(
\I[R],
[(\alpha_1,\mathrm{str}),\ldots,(\alpha_k,\mathrm{str}),
(\mathtt{bool},\mathrm{dom})],
\tau(\code{Bool})
\bigr).
\]
Boolean connectives on structural formulas are interpreted pointwise using the
same fuzzy operators as ordinary formulas. Thus,
\[
\I(\neg\rho)
=
\bigl(
N(\val{\I(\rho)}),
\axes(\I(\rho)),
\tau(\code{Bool})
\bigr).
\]
For \(\circ\in\{\land,\lor,\imp,\leftrightarrow\}\), the operands are first
aligned by name and role:
\[
\operatorname{align}
\bigl(
\I(\rho),\I(\sigma)
\bigr)
=
(\widehat{\I(\rho)},\widehat{\I(\sigma)}).
\]
Then
\[
\I(\rho\circ\sigma)
=
\bigl(
C_\circ
\bigl(
\val{\widehat{\I(\rho)}},
\val{\widehat{\I(\sigma)}}
\bigr),
\axes(\widehat{\I(\rho)}),
\tau(\code{Bool})
\bigr).
\]
The result is again a truth-valued annotated tensor, with structural axes given
by the aligned free structural axes of the input structural formulas.

\paragraph{Denotation of formulas.}
Formulas are interpreted by structural recursion over the formula grammar.
The symbols \(N\), \(C_\circ\), and \(A_Q\) denote, respectively, the fuzzy
negator, the fuzzy operator associated with the binary connective \(\circ\),
and the aggregation operator associated with
\(Q\in\{\forall,\exists\}\). These operators act on underlying tensor values;
the denotation rules below specify how the resulting values are equipped with
axis annotations and types.

\begin{enumerate}[label=(\roman*), itemsep=0pt]

\item \emph{Atomic formulas.}
An atomic formula
\[
p(q_1,\ldots,q_m)
\]
is interpreted analogously to a function application, except that its output
type is \(\tau(\code{Bool})\). Assume that each argument has already been
interpreted as an annotated tensor
\[
\I(q_j)=(v_j,\axes(q_j),\tau_j),
\qquad j=1,\ldots,m.
\]
Let
\[
\sort_{\mathrm{in}}(p)=\langle s_1,\ldots,s_m\rangle,
\]
and let
\[
\indim(p)=\Delta_{\mathrm{in}}
=
\langle d_1,\ldots,d_r\rangle,
\qquad
\outdim(p)=\Delta_{\mathrm{out}}
=
\langle e_1,\ldots,e_q\rangle.
\]

The argument denotations are first aligned by name and role on their free
variable and structural axes:
\[
\operatorname{align}
\bigl(
\I(q_1),\ldots,\I(q_m)
\bigr)
=
(\widehat q_1,\ldots,\widehat q_m).
\]
Among the aligned structural axes, those whose structural dimensions occur in
\(\Delta_{\mathrm{in}}\) are consumed by the predicate application. If an
argument carries such an axis, it contributes its indexed values along that
axis. If it does not carry such an axis, it is broadcast along that dimension.
The predicate grounding
\[
\I[p]:
\left(
\mathbb{R}^{|\tau_1|}
\times\cdots\times
\mathbb{R}^{|\tau_m|}
\right)^{\ell_{d_1}\times\cdots\times\ell_{d_r}}
\longrightarrow
[0,1]^{\ell_{e_1}\times\cdots\times\ell_{e_q}}
\]
is then applied to the aligned consumed structural axes and to the domain axes
of the arguments. Variable axes and structural axes not consumed by
\(\Delta_{\mathrm{in}}\) are external axes and are propagated pointwise.

\item \emph{Connectives.}
Logical connectives act pointwise on the underlying tensor values of
truth-valued annotated tensors. If
\[
\I(\phi)
=
(\val{\phi},\axes(\phi),\tau(\code{Bool})),
\]
then negation is interpreted as
\[
\I(\neg\phi)
=
\bigl(
N(\val{\phi}),
\axes(\phi),
\tau(\code{Bool})
\bigr).
\]

For a binary connective \(\circ \in \{\land,\lor,\imp,\leftrightarrow\},\)
the operand denotations are first aligned by name and role, then the connective is applied pointwise to their underlying tensor values:
\[
\I(\phi \circ \psi)
=
\bigl(
C_\circ
\bigl(
\val{\widehat{\I(\phi)}},
\val{\widehat{\I(\psi)}}
\bigr),
\axes(\widehat{\I(\phi)}),
\tau(\code{Bool})
\bigr).
\]
The resulting formula has type \(\tau(\code{Bool})\), carries the aligned free
variable and structural axes, and retains the Boolean domain axis.

\item \emph{First-order quantification.}
First-order quantifiers aggregate over variable axes. Suppose
\[
\I(\phi)
=
(\val{\phi},\axes(\phi),\tau(\code{Bool})),
\]
where \(x_1,\ldots,x_h\in\fv(\phi)\). Given a quantified formula
\[
Q\,x_1,\ldots,x_h:\phi,
\qquad
Q\in\{\forall,\exists\},
\]
we define
\[
\I(Q\,x_1,\ldots,x_h:\phi)
=
\bigl(
A_Q^{x_1,\ldots,x_h}(\val{\phi}),
\axes(\phi)\setminus
\bigl[
(x_1,\mathrm{var}),\ldots,(x_h,\mathrm{var})
\bigr],
\tau(\code{Bool})
\bigr).
\]
Here \(A_Q^{x_1,\ldots,x_h}\) denotes aggregation of the underlying tensor
value along the variable axes named \(x_1,\ldots,x_h\). These axes are removed
from the resulting annotated tensor; all other axes and the Boolean type are
preserved.

\item \emph{Structural quantification.}
Structural quantifiers aggregate over structural axes. Suppose
\[
\I(\phi)
=
(\val{\phi},\axes(\phi),\tau(\code{Bool})).
\]
If
\[
Q\,\alpha_1,\ldots,\alpha_h:\phi,
\qquad
Q\in\{\forall,\exists\},
\]
then
\[
\I(Q\,\alpha_1,\ldots,\alpha_h:\phi)
=
\bigl(
A_Q^{\alpha_1,\ldots,\alpha_h}(\val{\phi}),
\axes(\phi)\setminus
\bigl[
(\alpha_1,\mathrm{str}),\ldots,(\alpha_h,\mathrm{str})
\bigr],
\tau(\code{Bool})
\bigr).
\]
Here \(A_Q^{\alpha_1,\ldots,\alpha_h}\) aggregates the tensor value
along the structural axes named \(\alpha_1,\ldots,\alpha_h\). These axes are
removed from the result, while all unbound variable axes, remaining structural
axes, and the Boolean type are preserved.

\item \emph{Guarded quantification.}
Guarded quantification uses the value of the guard as a mask or weight for
aggregation. In the first-order case,
\[
Q\,x_1,\ldots,x_h \mid \psi : \phi,
\]
the body and guard denotations are first aligned:
\[
\operatorname{align}
\bigl(
\I(\phi),\I(\psi)
\bigr)
=
\bigl(
\widehat{\I(\phi)},\widehat{\I(\psi)}
\bigr).
\]
Here \(\widehat{\I(\phi)}\) and \(\widehat{\I(\psi)}\) are aligned annotated tensors. The denotation \(\I(Q\,x_1,\ldots,x_h \mid \psi :\phi)\)
is 
\[\bigl(
A_Q^{x_1,\ldots,x_h}
\bigl(
\val{\widehat{\I(\phi)}};
\val{\widehat{\I(\psi)}}
\bigr),
\axes(\widehat{\I(\phi)})\setminus
\bigl[
(x_1,\mathrm{var}),\ldots,(x_h,\mathrm{var})
\bigr],
\tau(\code{Bool})
\bigr).
\]
The aggregation is performed over the variable axes \(x_1,\ldots,x_h\), which are
removed from the resulting axis annotation.

For structural guarded quantification,
\[
Q\,\alpha_1,\ldots,\alpha_h \mid \rho : \phi,
\]
the body and structural guard denotations are first aligned:
\[
\operatorname{align}
\bigl(
\I(\phi),\I(\rho)
\bigr)
=
(\widehat{\I(\rho)},
\widehat{\I(\sigma)}
).
\]
Then \(\I(Q\,\alpha_1,\ldots,\alpha_h \mid \rho : \phi)\)
is defined as 
\[
\bigl(
A_Q^{\alpha_1,\ldots,\alpha_h}
\bigl(
\val{\widehat{\I(\phi)}};
\val{\widehat\rho}
\bigr),
\axes(\widehat{\I(\phi)})\setminus
\bigl[
(\alpha_1,\mathrm{str}),\ldots,(\alpha_h,\mathrm{str})
\bigr],
\tau(\code{Bool})
\bigr).
\]
The aggregation is performed over the structural axes
\(\alpha_1,\ldots,\alpha_h\), which are removed from the resulting axis
annotation.

\item \emph{Diagonal quantification.}
Diagonal quantification identifies several variable axes positionally before
aggregation. The parenthesized quantifier
\[
Q\,(x_1,\ldots,x_h):\phi
\]
is well defined only if the variable axes named \(x_1,\ldots,x_h\) in
\(\axes(\phi)\) have the same extent:
\[
b_{x_1}=\cdots=b_{x_h}.
\]
Suppose
\(
\I(\phi)
=
(\val{\phi},\axes(\phi),\tau(\code{Bool})).
\)
The denotation is obtained by first restricting \(\val{\phi}\) to the diagonal
subtensor \(\operatorname{diag}_{x_1,\ldots,x_h}(\val{\theta})\) on which the indices of \(x_1,\ldots,x_h\) coincide, and then
applying \(A_Q\) to the resulting shared diagonal axis:
\[
\bigl(
A_Q^{\Delta_{x_1,\ldots,x_h}}
\bigl(
\operatorname{diag}_{x_1,\ldots,x_h}(\val{\phi})
\bigr),
\axes(\I(\phi))\setminus
\bigl[
(x_1,\mathrm{var}),\ldots,(x_h,\mathrm{var})
\bigr],
\tau(\code{Bool})
\bigr).
\]
Thus, diagonal quantification ranges over aligned tuples rather than over the full Cartesian product of the quantified variables.

For the guarded form
\(Q\,(x_1,\ldots,x_h)\mid\psi:\phi,\)
a preliminary alignment of the denotations of the body and guard is performed.
\end{enumerate}
\section{Learning}
\label{s:learning}

The previous sections introduced the syntax of \sltn{}
(Section~\ref{s:sltn-logic}) and its semantics
(Section~\ref{s:semantics}), which assign to each well formed expression an
annotated tensor under an interpretation \(\mathcal{I}\). We now consider
\emph{learning}: the problem of choosing the parameters of an interpretation so
as to maximize the satisfaction of a knowledge base.

Recall from Section~\ref{s:grammar} that a knowledge base
\(\mathcal{K}\) is a finite set of clauses, i.e., closed formulas with no free
first-order or structural variables. Therefore, each clause
\(\phi\in\mathcal{K}\) is evaluated, under an interpretation
\(\mathcal{I}\), as a scalar fuzzy truth degree in \([0,1]\). Learning is then
driven by the satisfaction degrees of these clauses. This setting closely follows the standard LTN formulation
\citep{badreddine_logic_2022}. The main difference is that \sltn{} also allows
knowledge-base satisfaction to be treated explicitly as a multi-objective
optimization problem. We use the running example of
Examples~\ref{ex:video-signature} and~\ref{ex:video-formulas} to illustrate the
implementation.

In learning, some groundings assigned by the interpretation may depend on
real-valued parameters \(\theta\in\Theta\), where \(\Theta\) is a hypothesis
space. We write \(\mathcal{I}_{\theta}\) for such a parametric interpretation.
For a symbol \(a\) whose grounding is parameter-dependent, we write
\(\mathcal{I}_{\theta}[a]\) for its grounding. The denotation of an expression
\(e\) under the same interpretation is written \(\mathcal{I}_{\theta}(e)\).

\begin{definition}[Theory]
\label{def:theory}
An \sltn{} \emph{theory} is a triple
\[
\mathcal{T}
=
\langle \mathcal{K},\mathcal{I}_{\theta},\Theta\rangle,
\]
where \(\mathcal{K}\) is a knowledge base over a signature \(\Sigma\),
\(\mathcal{I}_{\theta}\) is a parametric interpretation of \(\Sigma\), and
\(\Theta\) is the hypothesis space of the parameters \(\theta\). An
\emph{interpreted theory}
\[
\langle \mathcal{K},\mathcal{I}_{\theta}\rangle
\]
is a theory equipped with a particular parameter value
\(\theta\in\Theta\).
\end{definition}

Let \(\mathcal{K}=\{\phi_1,\ldots,\phi_m\}\) and define the satisfaction degree of the \(i\)-th clause by
\[
s_i(\theta)
=
\mathcal{I}_{\theta}(\phi_i)
\in[0,1].
\]
Learning from \(\mathcal{K}\) amounts to making all clause satisfactions
\[
s(\theta)
=
\bigl(s_1(\theta),\ldots,s_m(\theta)\bigr)
\in[0,1]^m
\]
as high as possible. Thus, knowledge-base satisfaction can naturally be viewed
as a multi-objective optimization problem, where each clause contributes one
objective.

\paragraph{Scalarized learning.}
A standard approach is to reduce the vector of clause satisfactions to a single
training objective by means of a satisfaction aggregation operator
\(\SatAgg\). The satisfaction of \(\mathcal{K}\) under
\(\mathcal{I}_{\theta}\) is defined as
\[
\SatAgg_{\phi\in\mathcal{K}}
\mathcal{I}_{\theta}(\phi)
=
\SatAgg
\bigl(
s_1(\theta),\ldots,s_m(\theta)
\bigr)
\in[0,1],
\]
where \(\SatAgg\) is specified by the logic of the interpretation. Under this
scalarization, learning is formulated as
\begin{equation}
\label{eq:learning}
\theta^\ast
=
\argmax_{\theta\in\Theta}
\SatAgg_{\phi\in\mathcal{K}}
\mathcal{I}_{\theta}(\phi).
\end{equation}
Equivalently, one may minimize the scalar loss
\[
L_{\mathrm{scal}}(\theta)
=
1-
\SatAgg
\bigl(
s_1(\theta),\ldots,s_m(\theta)
\bigr).
\]

When the parameter-dependent groundings and the chosen fuzzy operators are
implemented by differentiable tensor operations, the objective in
Equation~\eqref{eq:learning} can be optimized using gradient-based methods.
Differentiability depends on the chosen interpretation, on the neural or tensor maps used as groundings, and on the selected fuzzy connectives and aggregation operators.

Scalarized learning is often effective: if all clauses are highly satisfied,
then the aggregated satisfaction is high. The converse, however, depends on the
aggregation operator. During optimization,
some clauses can be favored while others remain poorly
satisfied, especially in the presence of conflicting objectives. Choosing a non-compensatory or weakly compensatory \(\SatAgg\) can mitigate this effect by penalizing low clause satisfactions more strongly.

In \sltn{}, the satisfaction aggregator is specified in the
interpretation's logic. For example:
\begin{lstlisting}[language=Python]
logic["SatAgg"] = AggregPMeanError(p=2)
\end{lstlisting}
Further details on available scalar aggregators are given in
Appendix~\ref{a:operators}.

\paragraph{Multi-objective learning.}
Instead of aggregating clause satisfactions before differentiation, \sltn{} can
retain the per-clause losses
\[
L(\theta)
=
\bigl(
1-s_1(\theta),\ldots,1-s_m(\theta)
\bigr)
\in[0,1]^m
\]
and combine their gradients directly. Let
\[
J_{\theta}L(\theta)
\in
\mathbb{R}^{m\times|\theta|}
\]
be the Jacobian of the vector-valued loss, with one row per clause loss and
one column per learnable scalar parameter. A Jacobian-descent aggregator
\[
A:
\mathbb{R}^{m\times|\theta|}
\to
\mathbb{R}^{|\theta|}
\]
maps this Jacobian to a single update direction. The parameters are then
updated according to
\[
\theta
\leftarrow
\theta
-
\eta\,
A\bigl(J_{\theta}L(\theta)\bigr),
\]
where \(\eta>0\) is the learning rate.

Thus, scalarized learning combines clause satisfactions before
differentiation, whereas multi-objective learning differentiates clause losses
separately and combines their gradients afterwards. The satisfaction aggregator \(\SatAgg\) acts on truth degrees, while the Jacobian-descent aggregator \(A\) acts on gradients.

The implementation relies on the \code{torchjd} library
\citep{jacobian_descent}, which provides Jacobian-descent methods for
multi-objective optimization. In the examples below, we use PCGrad as the
Jacobian-descent aggregator. PCGrad addresses conflicting per-objective
gradients by projecting gradients so as to reduce pairwise conflicts. Other
Jacobian-descent aggregators can be used through the same interface; we refer
the reader to the \code{torchjd} documentation and to \citep{jacobian_descent} for further details.

\paragraph{Training interface.}
\sltn{} supports both scalarized and multi-objective learning through the
common training interface \code{kb\_backward}
(Section~\ref{s:impl}). In scalarized mode, the clauses of
\(\mathcal{K}\) are evaluated under the current interpretation
\(\mathcal{I}_{\theta}\), their satisfaction degrees are aggregated by
\(\SatAgg\), and the resulting scalar loss is backpropagated. In
multi-objective mode, the per-clause losses are kept separate, their Jacobian
with respect to the learnable parameters is computed, and a
Jacobian-descent aggregator combines the corresponding gradients.

Before training, the knowledge base is processed by
\code{kb\_describe(kb)}, which validates the clauses and returns a labelled
dictionary mapping clause identifiers to formulas:
\begin{lstlisting}[language=Python]
kb = sltn.KB(phi_1, ..., phi_n)
kb_dict = kb_describe(kb)
\end{lstlisting}

During training, each minibatch is bound to the variables of the interpretation
by updating their groundings. For the running example, the minibatch tensors
are assigned to \(x\) and \(y\). The function \code{kb\_backward} then
evaluates the clauses under the current interpretation, constructs either the
scalarized loss or the multi-objective training signal, and backpropagates into
the learnable parameters of the interpretation.

A typical training step is:
\begin{lstlisting}[language=Python]
optimizer = torch.optim.Adam(interp.parameters())

for batch in dataloader:
    optimizer.zero_grad()

    interp["x"] = batch["x"]
    interp["y"] = batch["y"]

    stats = kb_backward(optimizer, interp, kb_dict, aggregator="pcgrad")

    optimizer.step()
\end{lstlisting}

If no Jacobian-descent aggregator is provided, \code{kb\_backward} uses the
interpretation's \(\SatAgg\) and performs scalarized learning by aggregation.
If an aggregator such as \code{"pcgrad"} is provided, the function performs
multi-objective learning by combining per-clause gradients. In both cases,
\code{kb\_backward} returns per-clause satisfaction statistics, which can be
used to monitor the contribution of each clause during training. The method
\code{interp.parameters()} returns an iterator over the learnable parameters
of the interpretation.
\section{Implementation details}
\label{s:impl}

Most implementation aspects have already been illustrated in the running example. This
section therefore summarizes the organization of the library and highlights the main
implementation choices not discussed above.

\sltn{} is implemented in Python on top of PyTorch~\citep{paszke_pytorch_2019}. Its code follows the same syntax--semantics separation adopted in this paper: signatures and abstract syntax trees are constructed independently of tensor groundings, while an \code{Interpretation} subsequently associates symbols with concrete tensors, PyTorch
modules, and fuzzy operators. Concretely, the library is organized into four main
packages,
\[
\code{sltn.signature}\rightarrow
\code{sltn.fol}\rightarrow
\code{sltn.parse}\rightarrow
\code{sltn.interpretation},
\]
which reflect the progression from signatures, to terms and formulas, to their
interpretation. The dependencies are designed to follow this order as far as possible. A limited interaction between parsing and signature-level constructs is nevertheless needed to support named definitions.

\paragraph{From signature to abstract syntax.}
A \code{Signature} is a validated registry of the typed symbols of
Section~\ref{s:signature}: each declaration method checks arities and reserved names and
returns a \code{Symbol} object. Parsing a formula string produces an explicit
abstract syntax tree that mirrors the grammar of Section~\ref{s:grammar}. Each syntax
node exposes its free first-order variables and free structural variables
compositionally. 

\paragraph{Interpretation and evaluation.}
An \code{Interpretation} binds the symbols of a signature to concrete groundings by
dictionary-style assignment. When applied to a parsed term or formula, it evaluates it
into the annotated-tensor form of Section~\ref{s:semantics} by structural recursion over
the syntax tree. Annotated tensors correspond to the \code{sltn.Tensor} class. Fuzzy connectives and aggregation operators are collected in a separate
\code{Logic} object.

Alignment and broadcasting are implemented by relying on standard PyTorch tensor
operations. In particular, the \(\operatorname{align}\) operation of
Section~\ref{s:semantics} inserts singleton axes where needed and then delegates the
pointwise operation to PyTorch's native broadcasting mechanism. This avoids explicitly
materializing operands to a common shape and is therefore important when structural axes
are large, for example in high-arity relations.

The implementation also follows PyTorch's device model. An interpretation can be created
on a target device by calling \code{to(device)} on the signature before constructing
the interpretation:
\begin{lstlisting}[language=Python]
device = torch.device("cuda")
interp = Interpretation(sig).to(device)
\end{lstlisting}
All objects declared in the signature are then automatically moved to the selected device
when they are grounded in the interpretation. Notice that, as in standard PyTorch, moving an object to
between devices may create a copy of the original object.

The interpretation interface also provides intuitive support for batching. In particular, variable groundings can be reassigned between training steps by dictionary-style updates:
\begin{lstlisting}[language=Python]
interp["x"] = batch["x"]
interp["y"] = batch["y"]
\end{lstlisting}
Thus, the same parsed formulas and knowledge base can be evaluated on different minibatches simply by updating the corresponding variable groundings in the
interpretation.

The library, together with tutorials, the running example and further example notebooks, is available at available at \url{https://github.com/logictensornetworks/sltn}.
\section{Related Work}
\label{s:relwork}

\sltn{} inherits its position in the neurosymbolic landscape from \ltn{}
\citep{badreddine_logic_2022}, which falls in the class of architectures
that integrate inductive and deductive reasoning by encoding the
satisfiability of logical knowledge directly into the loss function of a
neural network, alongside systems such as Semantic Loss
\citep{xu_semantic_2018}, \textsc{Lyrics} \citep{marra2019lyrics}, and the
teacher-student framework of \citet{Harnessing}. This differs from
approaches that instead modify the predictions of a base network through a
separate constraint-satisfaction layer (Deep Logic Models
\citep{DBLP:conf/pkdd/MarraGDG19a}, KENN \citep{daniele2019knowledge}), or
that perform (probabilistic) logical inference on top of a base network's
predictions (DeepProbLog \citep{DeepProbLog}). We refer to
\citet{badreddine_logic_2022} for a detailed comparison along these lines;
the contribution of this paper is orthogonal to that categorization, since
it concerns the \emph{expressiveness of the grounded language} rather than
the point in the training loop where logic and learning are coupled.

\paragraph{Structured and relational extensions of neurosymbolic
frameworks.} The need to reason over sequential or relational structure
that is not a plain first-order domain is not unique to \ltn{}. Lifted
relational neural networks \citep{sourek2018lifted} use a Datalog program
to compactly specify graph neural network architectures, with node/edge
structure playing a role similar to \sltn{}'s dimensions and structural
relations, but as a specification of the network's connectivity rather
than as part of the grounded logical language itself — in \sltn{}, a
structural relation such as $\mathrm{next}$ is a first-class term of the
formula, quantified and combined with ordinary connectives like any other
predicate. Statistical relational learning frameworks such as Markov Logic
Networks \citep{MLN} associate weighted first-order formulas with a
Markov random field over ground atoms, and can express positional
structure (e.g.\ a temporal successor relation) as an ordinary binary
predicate on individuals; \sltn{}'s structural relations instead live in a
separate namespace from sorts and their individuals specifically so that
axis alignment between formulas remains unambiguous
(Section~\ref{s:semantics}) and so that quantifying a large temporal or
positional axis does not require materializing it as a first-order
variable with its own individuals and their features. Weighted Real Logic
as used in Logical Neural Networks \citep{riegel2020logical} shares
\ltn{}'s idea of grounding connectives as differentiable operators inside
a modular, tree-shaped network, but — like the original \ltn{} — does not
single out a positional/structural axis role distinct from ordinary
variables.

\paragraph{Explicit formula languages.} \sltn{}'s parser and \code{Signature}
registry (Section~\ref{s:impl}) address a more implementation-level gap:
\code{LTNtorch} and the original \ltn{} have no string syntax for
formulas — knowledge bases are hand-composed by calling Python operator
overloads on \code{LTNObject}s — which makes arity, sort, and dimension
mismatches only detectable at grounding time (or not at all, if the shapes
happen to broadcast). By contrast, statistical relational learning and
probabilistic logic programming systems (e.g.\ Markov Logic Networks
\citep{MLN}, ProbLog \citep{DeepProbLog}) typically parse an explicit
logic-program syntax and validate a signature ahead of grounding; \sltn{}
brings a comparable syntax/semantics separation and static validation
(clause closedness, reserved names, arity checking) to the \ltn{} setting
while keeping its differentiable, tensor-native grounding.

\section{Conclusions and Future Work}
\label{s:concl}

Many domains relevant to neuro-symbolic reasoning involve data with inherent
structure, such as temporal order, sequential dependencies, spatial
organization, graph-like connectivity, and relations among positions. This
paper introduced \sltn{}, a neuro-symbolic framework that extends
\ltn{} by incorporating such structure directly into differentiable logical
reasoning. The proposed framework treats structural dimensions, structural
variables, and structural relations as explicit components of the logical
language, thereby enabling formulas to refer not only to individuals but also
to their organization within a structured domain.

The paper developed the main components of the framework: a typed signature
language for declaring structured symbols, a formula language with structural
annotation, selection, relations, and quantification, and a fuzzy tensor
semantics in which expressions are interpreted as annotated tensors with named
variable, structural, and domain axes. This semantics integrates structural
constructs with fuzzy connectives, quantifier aggregation, and differentiable
groundings, while preserving compatibility with the standard unstructured
\ltn{} setting. The framework is implemented as a modular Python library on
top of PyTorch, organized around explicit \code{Signature}, parser,
\code{Interpretation}, and learning components. The paper is intended to serve
as a companion to the library implementation, available at
\url{https://github.com/logictensornetworks/sltn}.

As an initial framework and software library, \sltn{} requires further
stabilization, testing, and systematic empirical validation beyond the
illustrative examples considered so far. Future work may proceed in several
complementary directions. First, dedicated structured theories should be
developed for concrete application domains, in which domain-specific objects,
functions, predicates, dimensions, and structural relations are introduced as
first-class \sltn{} constructs. Examples of such theories are currently in
preparation. Second, the expressive power of the language could be extended
with a small set of finitely recursive term-level computational primitives,
allowing simple structured computations to be expressed within formulas while
preserving differentiability and a well-defined tensor semantics. Third,
principled schedules for the power-mean exponent \(p\) should be investigated
in order to improve training stability and support more systematic
hyperparameter selection. Further directions include benchmarking of library components and a more detailed comparison with alternative
approaches to relational, temporal, spatial, and graph-based learning.

\bibliographystyle{plain}
\bibliography{biblio}

@inproceedings{DBLP:conf/pkdd/MarraGDG19a,
  author    = {Giuseppe Marra and
               Francesco Giannini and
               Michelangelo Diligenti and
               Marco Gori},
  title     = {Integrating Learning and Reasoning with Deep Logic Models},
  booktitle = {Machine Learning and Knowledge Discovery in Databases - European Conference,
               {ECML} {PKDD} 2019, W{\"{u}}rzburg, Germany, September 16-20,
               2019, Proceedings, Part {II}},
  series    = {Lecture Notes in Computer Science},
  volume    = {11907},
  pages     = {517--532},
  publisher = {Springer},
  year      = {2019},
  url       = {https://doi.org/10.1007/978-3-030-46147-8\_31},
  doi       = {10.1007/978-3-030-46147-8\_31},
  bibsource = {dblp computer science bibliography, https://dblp.org}
}

@article{riegel2020logical,
	title = {Logical {Neural} {Networks}},
	url = {http://arxiv.org/abs/2006.13155},
	urldate = {2021-05-17},
	journal = {arXiv:2006.13155 [cs]},
	author = {Riegel, Ryan and Gray, Alexander and Luus, Francois and Khan, Naweed and Makondo, Ndivhuwo and Akhalwaya, Ismail Yunus and Qian, Haifeng and Fagin, Ronald and Barahona, Francisco and Sharma, Udit and Ikbal, Shajith and Karanam, Hima and Neelam, Sumit and Likhyani, Ankita and Srivastava, Santosh},
	month = jun,
	year = {2020},
	note = {arXiv: 2006.13155},
}

@article{MLN,
 author = {Richardson, Matthew and Domingos, Pedro},
 title = {Markov Logic Networks},
 journal = {Mach. Learn.},
 issue_date = {February  2006},
 volume = {62},
 number = {1-2},
 month = feb,
 year = {2006},
 issn = {0885-6125},
 pages = {107--136},
 numpages = {30},
 url = {http://dx.doi.org/10.1007/s10994-006-5833-1},
 doi = {10.1007/s10994-006-5833-1},
 acmid = {1113910},
 publisher = {Kluwer Academic Publishers},
 address = {Hingham, MA, USA},
}

@inproceedings{LTNIJCAI,
  author    = {Ivan Donadello and Luciano Serafini and Artur d'Avila Garcez},
  title     = {Logic Tensor Networks for Semantic Image Interpretation},
  booktitle = {Proceedings of the Twenty-Sixth International Joint Conference on
               Artificial Intelligence, {IJCAI-17}},
  pages     = {1596--1602},
  year      = {2017},
  doi       = {10.24963/ijcai.2017/221},
  url       = {https://doi.org/10.24963/ijcai.2017/221},
}

@article{LTN,
  author    = {Luciano Serafini and
               Artur d'Avila Garcez},
  title     = {Logic Tensor Networks: Deep Learning and Logical Reasoning from Data
               and Knowledge},
  journal   = {CoRR},
  volume    = {abs/1606.04422},
  year      = {2016},
  url       = {http://arxiv.org/abs/1606.04422},
  archivePrefix = {arXiv},
  eprint    = {1606.04422},
  bibsource = {dblp computer science bibliography, https://dblp.org}
}

@inproceedings{Harnessing,
    title = "Harnessing Deep Neural Networks with Logic Rules",
    author = "Hu, Zhiting  and
      Ma, Xuezhe  and 
      Liu, Zhengzhong  and
      Hovy, Eduard  and
      Xing, Eric",
    booktitle = "Proceedings of the 54th Annual Meeting of the Association for Computational Linguistics (Volume 1: Long Papers)",
    month = aug,
    year = "2016",
    address = "Berlin, Germany",
    publisher = "Association for Computational Linguistics",
    doi = "10.18653/v1/P16-1228",
    pages = "2410--2420",
}

@inproceedings{DeepProbLog,
 author = {Manhaeve, Robin and Dumancic, Sebastijan and Kimmig, Angelika and Demeester, Thomas and Raedt, Luc De},
 title = {DeepProbLog: Neural Probabilistic Logic Programming},
 booktitle = {Proceedings of the 32nd International Conference on Neural Information Processing Systems},
 series = {NeurIPS'18},
 year = {2018},
 location = {Montr\&\#233;al, Canada},
 pages = {3753--3763},
 numpages = {11},
 url = {http://dl.acm.org/citation.cfm?id=3327144.3327291},
 acmid = {3327291},
 publisher = {Curran Associates Inc.},
 address = {USA}
}

@article{van_krieken_analyzing_2020,
	title = {Analyzing {Differentiable} {Fuzzy} {Logic} {Operators}},
	url = {http://arxiv.org/abs/2002.06100},
	urldate = {2020-05-07},
	journal = {arXiv:2002.06100 [cs]},
	author = {van Krieken, Emile and Acar, Erman and van Harmelen, Frank},
	month = feb,
	year = {2020},
	note = {arXiv: 2002.06100}
}

@inproceedings{daniele2019knowledge,
  title={Knowledge Enhanced Neural Networks},
  author={Daniele, Alessandro and Serafini, Luciano},
  booktitle={Pacific Rim International Conference on Artificial Intelligence},
  pages={542--554},
  year={2019},
  organization={Springer}
}

@article{sourek2018lifted,
  title={Lifted relational neural networks: Efficient learning of latent relational structures},
  author={Sourek, Gustav and Aschenbrenner, Vojtech and Zelezny, Filip and Schockaert, Steven and Kuzelka, Ondrej},
  journal={Journal of Artificial Intelligence Research},
  volume={62},
  pages={69--100},
  year={2018}
}

@inproceedings{marra2019lyrics,
  title={LYRICS: A General Interface Layer to Integrate Logic Inference and Deep Learning},
  author={Marra, Giuseppe and Giannini, Francesco and Diligenti, Michelangelo and Gori, Marco},
  booktitle={Joint European Conference on Machine Learning and Knowledge Discovery in Databases},
  pages={283--298},
  year={2019},
  organization={Springer}
}

@inproceedings{serafini2016learning,
  title={Learning and reasoning with logic tensor networks},
  author={Serafini, Luciano and Garcez, Artur d’Avila},
  booktitle={Conference of the Italian Association for Artificial Intelligence},
  pages={334--348},
  year={2016},
  organization={Springer}
}

@inproceedings{donadello2019compensating,
  title={Compensating supervision incompleteness with prior knowledge in semantic image interpretation},
  author={Donadello, Ivan and Serafini, Luciano},
  booktitle={2019 International Joint Conference on Neural Networks (IJCNN)},
  pages={1--8},
  year={2019},
  organization={IEEE}
}

@inproceedings{xu_semantic_2018,
	title = {A {Semantic} {Loss} {Function} for {Deep} {Learning} with {Symbolic} {Knowledge}},
	url = {http://proceedings.mlr.press/v80/xu18h.html},
	language = {en},
	urldate = {2021-08-04},
	booktitle = {International {Conference} on {Machine} {Learning}},
	publisher = {PMLR},
	author = {Xu, Jingyi and Zhang, Zilu and Friedman, Tal and Liang, Yitao and Broeck, Guy},
	month = jul,
	year = {2018},
	note = {ISSN: 2640-3498},
	pages = {5502--5511},
}

@article{badreddine_logic_2022,
  title   = {Logic Tensor Networks},
  author  = {Badreddine, Samy and d'Avila Garcez, Artur and Serafini, Luciano and Spranger, Michael},
  journal = {Artificial Intelligence},
  volume  = {303},
  pages   = {103649},
  year    = {2022},
  publisher = {Elsevier}
}

@misc{paszke_pytorch_2019,
  title   = {PyTorch: An Imperative Style, High-Performance Deep Learning Library},
  author  = {Paszke, Adam and others},
  year    = {2019},
  eprint  = {1912.01703},
  archivePrefix = {arXiv}
}

@article{bouaziz2025enhancing,
  author  = {Bouaziz, Youssef and Barra, Vincent},
  title   = {Enhancing Multi-Label Learning in Visual Scenes with Logic Tensor Networks and Positive-Unlabeled Learning},
  year    = {2025},
  url     = {https://ssrn.com/abstract=5458386},
  doi     = {10.2139/ssrn.5458386},
  urldate = {2026-07-30},
  journal    = {SSRN Electronic Journal}
}

@article{manigrasso2026boosting,
  title={Boosting zero-shot learning through neuro-symbolic integration},
  author={Manigrasso, Francesco and Lamberti, Fabrizio and Morra, Lia},
  journal={Pattern Recognition},
  volume={170},
  pages={111869},
  year={2026},
  publisher={Elsevier}
}

@article{colamonaco2026weakly,
  title={Weakly Supervised Segmentation as Semantic-Based Regularization},
  author={Colamonaco, Stefano and Florea, Andrei-Bogdan and Maene, Jaron},
  journal={arXiv preprint arXiv:2605.13674},
  year={2026}
}

@article{boscarato2026first,
  title={First-Order Temporal Logic Tensor Networks},
  author={Boscarato, Luca and Donadello, Ivan and Artale, Alessandro and Montali, Marco and Maggi, Fabrizio Maria},
  journal={arXiv preprint arXiv:2606.29972},
  year={2026}
}

@inproceedings{djenouri2024neurosymbolic,
  title={Neurosymbolic Visual Transform Based on Logic Tensor Network for Defect Detection},
  author={Djenouri, Youcef and Belbachir, Ahmed Nabil and Belhadi, Asma and Michalak, Tomasz},
  booktitle={European Conference on Computer Vision},
  pages={18--34},
  year={2024},
  organization={Springer}
}

@article{gang2026feedltn,
AUTHOR = {Yang, Gang and Ni, Lin and Geng, Junfeng and Peng, Xiang},
TITLE = {FedLTN-CubeSat: Neuro-Symbolic Federated Learning for Intrusion Detection in LEO CubeSat Constellations},
JOURNAL = {Mathematics},
VOLUME = {14},
YEAR = {2026},
NUMBER = {6},
ARTICLE-NUMBER = {1047},
URL = {https://www.mdpi.com/2227-7390/14/6/1047},
ISSN = {2227-7390},
DOI = {10.3390/math14061047}
}

@inproceedings{carraro2023overcoming,
  title={Overcoming recommendation limitations with neuro-symbolic integration},
  author={Carraro, Tommaso},
  booktitle={Proceedings of the 17th ACM Conference on Recommender Systems},
  pages={1325--1331},
  year={2023}
}

@inproceedings{eckert2026extending,
  title={Extending Logic Tensor Networks to Implicit Feedback for Representation-Aware Music Recommendation},
  author={Eckert, Hannah and Lesota, Oleg and Schedl, Markus},
  booktitle={European Conference on Information Retrieval},
  pages={428--441},
  year={2026},
  organization={Springer}
}

@inproceedings{carraro2024mitigating,
  title={Mitigating data sparsity via neuro-symbolic knowledge transfer},
  author={Carraro, Tommaso and Daniele, Alessandro and Aiolli, Fabio and Serafini, Luciano},
  booktitle={European Conference on Information Retrieval},
  pages={226--242},
  year={2024},
  organization={Springer}
}

@article{tan2024ontomedrec,
  title={OntoMedRec: Logically-pretrained model-agnostic ontology encoders for medication recommendation},
  author={Tan, Weicong and Wang, Weiqing and Zhou, Xin and Buntine, Wray and Bingham, Gordon and Yin, Hongzhi},
  journal={World Wide Web},
  volume={27},
  number={3},
  pages={28},
  year={2024},
  publisher={Springer}
}

@article{jacobian_descent,
  title={Jacobian Descent For Multi-Objective Optimization},
  author={Quinton, Pierre and Rey, Valérian},
  journal={arXiv preprint arXiv:2406.16232},
  year={2024}
}

@article{carraro2024ltntorch,
  title={LTNtorch: PyTorch implementation of Logic Tensor Networks},
  author={Carraro, Tommaso and Serafini, Luciano and Aiolli, Fabio},
  journal={arXiv preprint arXiv:2409.16045},
  year={2024}
}

@inproceedings{de2024enhancing,
  title={Enhancing Logical Tensor Networks: Integrating Uninorm-Based Fuzzy Operators for Complex Reasoning},
  author={de Campos Souza, Paulo Vitor and Apriceno, Gianluca and Dragoni, Mauro},
  booktitle={International Conference on Neural-Symbolic Learning and Reasoning},
  pages={68--79},
  year={2024},
  organization={Springer}
}

@article{mondal2025logic,
  title={A Logic Tensor Network-Based Neurosymbolic Framework for Explainable Diabetes Prediction},
  author={Mondal, Semanto and Ferraro, Antonino and Pecorelli, Fabiano and De Pietro, Giuseppe},
  journal={Applied Sciences},
  volume={15},
  number={21},
  pages={11806},
  year={2025},
  publisher={MDPI}
}

@article{gao2025enhancing,
  title={Enhancing Transcription Factor Prediction via Domain Knowledge Integration with Logic Tensor Networks},
  author={Gao, Liyuan and Sun, Linpeng and Sheng, Victor S},
  journal={IEEE Transactions on Computational Biology and Bioinformatics},
  year={2025},
  publisher={IEEE}
}

@article{MOSER_kernels,
title = {On the T-transitivity of kernels},
journal = {Fuzzy Sets and Systems},
volume = {157},
number = {13},
pages = {1787-1796},
year = {2006},
issn = {0165-0114},
doi = {https://doi.org/10.1016/j.fss.2006.01.007},
url = {https://www.sciencedirect.com/science/article/pii/S0165011406000078},
author = {Bernhard Moser},
}

@book{bullen2013handbook,
  title={Handbook of means and their inequalities},
  author={Bullen, Peter S},
  volume={560},
  year={2013},
  publisher={Springer Science \& Business Media}
  }

@inproceedings{bergamin2025integrating,
  title={Integrating Background Knowledge in Medical Semantic Segmentation with Logic Tensor Networks},
  author={Bergamin, Luca and Dimitri, Giovanna Maria and Aiolli, Fabio},
  booktitle={2025 International Joint Conference on Neural Networks (IJCNN)},
  pages={1--7},
  year={2025},
  organization={IEEE}
}

@inproceedings{de2026neuro,
  title={Neuro-symbolic learning for predictive process monitoring via two-stage logic tensor networks with rule pruning},
  author={De Santis, Fabrizio and Park, Gyunam and Zanichelli, Francesco},
  booktitle={Pacific-Asia Conference on Knowledge Discovery and Data Mining},
  pages={104--118},
  year={2026},
  organization={Springer}
}

@inproceedings{gaikwad12026neuro,
  title={Neuro-Symbolic Process Anomaly Detection},
  author={Gaikwad$^1$, Devashish and van der Aalst$^1$, Wil MP and Park, Gyunam},
  booktitle={Advanced Information Systems Engineering: 38th International Conference, CAiSE 2026, Verona, Italy, June 8--12, 2026, Proceedings, Part II},
  pages={240},
  year={2026},
  organization={Springer Nature}
}

@inproceedings{umili2023grounding,
  title={Grounding ltlf specifications in image sequences},
  author={Umili, Elena and Capobianco, Roberto and De Giacomo, Giuseppe},
  booktitle={Proceedings of the International Conference on Principles of Knowledge Representation and Reasoning},
  pages={668--678},
  year={2023}
}

@article{dai2025large,
  title={Large language model enhanced logic tensor network for stance detection},
  author={Dai, Genan and Liao, Jiayu and Zhao, Sicheng and Fu, Xianghua and Peng, Xiaojiang and Huang, Hu and Zhang, Bowen},
  journal={Neural Networks},
  volume={183},
  pages={106956},
  year={2025},
  publisher={Elsevier}
}

@inproceedings{manigrasso2024probing,
  title={Probing llms for logical reasoning},
  author={Manigrasso, Francesco and Schouten, Stefan and Morra, Lia and Bloem, Peter},
  booktitle={International conference on neural-symbolic learning and reasoning},
  pages={257--278},
  year={2024},
  organization={Springer}
}

@inproceedings{haufe2026large,
  title={From Large Language Model Predicates to Logic Tensor Networks: Neurosymbolic Offer Validation in Regulated Procurement},
  author={Haufe, Cedric S and Stolzenburg, Frieder},
  booktitle={German Conference on Artificial Intelligence (K{\"u}nstliche Intelligenz)},
  pages={236--243},
  year={2026},
  organization={Springer}
}

@article{yang2025neuro,
  title={Neuro-symbolic artificial intelligence: Towards improving the reasoning abilities of large language models},
  author={Yang, Xiao-Wen and Shao, Jie-Jing and Guo, Lan-Zhe and Zhang, Bo-Wen and Zhou, Zhi and Jia, Lin-Han and Dai, Wang-Zhou and Li, Yu-Feng},
  journal={arXiv preprint arXiv:2508.13678},
  year={2025}
}

@article{upreti2026logic,
  title={Logic Tensor Network-Enhanced Generative Adversarial Network},
  author={Upreti, Nijesh and Belle, Vaishak},
  journal={arXiv preprint arXiv:2601.03839},
  year={2026}
}

@article{squareplus,
  author       = {Jonathan T. Barron},
  title        = {Squareplus: {A} Softplus-Like Algebraic Rectifier},
  journal      = {CoRR},
  volume       = {abs/2112.11687},
  year         = 2021,
  url          = {https://arxiv.org/abs/2112.11687},
  eprinttype   = {arXiv},
  eprint       = {2112.11687},
  bibsource    = {dblp computer science bibliography, https://dblp.org}
}

\appendix

\section{Fuzzy Operators and Their Implementation in \sltn{}}
\label{a:operators}

This appendix formalizes the fuzzy operators implemented in \sltn{} and relates
them to standard constructions from fuzzy logic. The presentation follows the
operator analysis commonly used in Logic Tensor Networks
\citep{badreddine_logic_2022}, adapted to the operators, aggregators, and implementation choices supported by \sltn{}. We also discuss here numerical and optimization-related issues that arise when these operators are used inside differentiable learning pipelines.

In the implementation, a \code{Logic} object specifies how each syntactic
logical construct is interpreted numerically. It assigns operator instances to
negation, conjunction, disjunction, implication, equivalence, universal and
existential aggregation, and knowledge-base satisfaction aggregation. These
operators act on tensor values in \([0,1]\) and are applied pointwise, except
for aggregators, which reduce one or more tensor axes. Structural quantification
uses the same aggregation mechanisms as first-order quantification; the
difference lies in the axes over which aggregation is performed, as described
in Section~\ref{s:semantics}.

\subsection{Preliminaries in fuzzy logic}

We first recall the algebraic properties of the connectives used in \sltn{}. For each connective, we give the corresponding fuzzy-logic definition and then identify the implementation classes provided by the framework.

\paragraph{Negation.} A \emph{fuzzy negation} is a function $N:[0,1]\to[0,1]$ satisfying
\[
N(0)=1, \qquad N(1)=0,
\]
and monotonicity in the decreasing direction:
\[
x\le y \quad \Rightarrow \quad N(x)\ge N(y).
\]
A negation is \emph{strict} if it is continuous and strictly decreasing, and \emph{strong} if it is involutive:
\[
N(N(x))=x
\qquad
\text{for all } x\in[0,1].
\]

\noindent\sltn{} implements the negation operators:
\begin{itemize}[itemsep=0pt]
\item
$\code{NotStandard}(a)=1-a$. This is the default negation.
\item
$\code{NotGodel}(a)=\mathbbm{1}[a=0].$ G\"odel negation.
\end{itemize}

\paragraph{Conjunction.}
A \emph{fuzzy conjunction} is a function $C:[0,1]^2\to[0,1]$ satisfying the Boolean boundary conditions
\[
C(0,0)=C(0,1)=C(1,0)=0,
\qquad
C(1,1)=1,
\]
monotonicity in both arguments and commutativity. In fuzzy logic, conjunctions are commonly modeled by t-norms. A t-norm $T:[0,1]^2\to[0,1]$ additionally satisfies associativity and has $1$ as neutral element
\[
T(x,1)=x,
\]
for all \(x\in[0,1]\). 

\noindent
\sltn{} implements the following conjunction operators:
\begin{itemize}[itemsep=0pt]
    \item
    $\code{AndMin}(a,b)=\min(a,b)$. 
    Gödel conjunction (t-norm).
    \item
    $\code{AndProd}(a,b)=ab$. 
    Goguen conjunction (t-norm).

    \item
    $\code{AndLuk}(a,b)=\max(a+b-1,0)$. 
    Łukasiewicz conjunction (t-norm).

    \item
    $\code{AndCos}(a,b)=
    ab-\sqrt{1-a^2}\sqrt{1-b^2}$.
    This is a cosine-based t-norm introduced by \citep{MOSER_kernels} in the study of transitivity properties of kernels in a fuzzy-logic setting.

    \item
    $\code{AndPMean}(a,b)=
    \left(\frac{a^p+b^p}{2}\right)^{1/p}$.
    This is a generalized-mean operator. For positive inputs, it converges to the geometric mean as $p\to0$:
    \[
    \lim_{p\to0}
    \left(\frac{a^p+b^p}{2}\right)^{1/p}
    =
    \sqrt{ab}.
    \]
    This operator is not a t-norm, since it is not associative and does not have a neutral element.
    It is approximately a conjunction when $p\to 0$. It can be useful in differentiable settings because nested product conjunctions may yield vanishing gradients when several conjuncts are close to zero.
\end{itemize}

\paragraph{Disjunction.}
Dually, a \emph{fuzzy disjunction} is a function $D:[0,1]^2\to[0,1]$ satisfying
\[
D(0,0)=0,
\qquad
D(0,1)=D(1,0)=D(1,1)=1,
\]
and monotonicity in both arguments. Disjunctions are commonly modeled by t-conorms. A t-conorm $S:[0,1]^2\to[0,1]$ additionally satisfies commutativity, associativity, monotonicity, and has $0$ as neutral element:
\[
S(x,0)=x,
\qquad
\text{for all } x\in[0,1].
\]
Given a t-norm $T$ and a negation $N$, the $N$-dual t-conorm $S$ is defined by
\begin{equation}
\label{eq:demorgan1}
S(a,b)=N\bigl(T(N(a),N(b))\bigr).
\end{equation}
If $N$ is strong, the converse relation also holds:
\begin{equation}
\label{eq:demorgan2}
T(a,b)=N\bigl(S(N(a),N(b))\bigr).
\end{equation}
In \sltn{}, the class \(\code{OrDual}_{C,N}\) implements Eq.~\eqref{eq:demorgan1}. Thus, it derives a disjunction from a configured conjunction $C$ and negation $N$. \sltn{} implements the following disjunction operators:
\begin{itemize}[itemsep=0pt]
    \item
    $\code{OrMax}(a,b)=\max(a,b)$.
    This is the G\"odel disjunction. It is the \code{NotStandard}-dual of \code{AndMin}:
    \[
    \max(a,b)=1-\min(1-a,1-b).
    \]

    \item
    $\code{OrProbSum}(a,b)=a+b-ab$.
    This is the Product/Goguen disjunction, also known as the probabilistic sum. It is the \code{NotStandard}-dual of \code{AndProd}:
    \[
    a+b-ab
    =
    1-(1-a)(1-b).
    \]
    \item
    $\code{OrLuk}(a,b)=\min(a+b,1)$.
    This is the \L{}ukasiewicz disjunction. It is the \code{NotStandard}-dual of \code{AndLuk}:
    \[
    \min(a+b,1)
    =
    1-\max((1-a)+(1-b)-1,0).
    \]
\end{itemize}

\paragraph{Implication.}
A \emph{fuzzy implication} is a function $I:[0,1]^2\to[0,1]$ satisfying the Boolean boundary conditions
\[
I(0,0)=I(0,1)=I(1,1)=1,
\qquad
I(1,0)=0.
\]
\sltn{} implements implications either directly or through two standard constructions: strong implication and residuated implication.
A strong implication, or simply \emph{S-implication}, generalizes material implication:
\begin{equation}
\label{eq:s-implication}
I_S(a,b)=S(N(a),b),
\end{equation}
where $N$ is a fuzzy negation and $S$ is a fuzzy disjunction. If $S$ is the $N$-dual of a t-norm $T$, Eq.~\eqref{eq:s-implication} can equivalently be written as
\begin{equation}
\label{eq:s-implication-dual}
I_S(a,b)=N\bigl(T(a,N(b))\bigr).
\end{equation}
In \sltn{}, $\code{ImpliesDual}_{C,N}$ implements this generic S-implication construction from a configured conjunction $C$ and negation $N$.

A residuated implication, or \emph{R-implication}, is induced by a t-norm $T$:
\begin{equation}
\label{eq:r-implication}
I_R(a,b)
=
\sup\{z\in[0,1]\mid T(a,z)\le b\}.
\end{equation}
This construction returns the greatest truth value that can be conjoined with the antecedent without exceeding the consequent. This construction need not admit a simple closed form for an arbitrary t-norm, but it does for the standard families listed below.

\sltn{} implements the following implication operators:
\begin{itemize}[itemsep=0pt]
    \item
    $\code{ImpliesKleeneDienes}(a,b)=\max(1-a,b)$.
    This is the S-implication obtained from \code{NotStandard} and \code{AndMin}. 
    \item
    $\code{ImpliesReichenbach}(a,b)=1-a+ab$.
    This is the S-implication obtained from \code{NotStandard} and \code{AndProd}. 
    \item
    $\code{ImpliesLuk}(a,b)=\min(1-a+b,1)$.
    This is both an S-implication and an R-implication. As an S-implication, it is obtained from \code{NotStandard} and \code{AndLuk}. 
    As an R-implication, it is the residuum of the \L{}ukasiewicz t-norm \code{AndLuk}.

    \item
    $\code{ImpliesGodel}(a,b)=
    \begin{cases}
    1, & a\le b,\\
    b, & \text{otherwise}.
    \end{cases}$
    This is the R-implication obtained from  \code{AndMin}.
    \item
    $\code{ImpliesGoguen}(a,b)=
    \begin{cases}
    1, & a\le b,\\
    b/a, & \text{otherwise}.
    \end{cases}$
    This is the R-implication, obtained from \code{AndProd}.
\end{itemize}

\paragraph{Equivalence.}
A \emph{fuzzy equivalence} is a function $E:[0,1]^2\to[0,1]$ that generalizes the Boolean biconditional. It satisfies the Boolean boundary conditions
\[
E(0,0)=E(1,1)=1,
\qquad
E(0,1)=E(1,0)=0,
\]
symmetry,
\[
E(a,b)=E(b,a),
\]
and reflexivity,
\[
E(a,a)=1.
\]
For each fixed $a$, $E(a,b)$ is non-decreasing for $b\le a$ and non-increasing for $b\ge a$. Thus, the maximum value is attained when $a=b$. In fuzzy logic, equivalence is often defined compositionally as mutual implication:
\begin{equation}
\label{eq:equiv-biconditional}
E(a,b)
=
C\bigl(I(a,b),I(b,a)\bigr),
\end{equation}
where $I$ is a fuzzy implication and $C$ is a fuzzy conjunction.

\noindent\sltn{} implements the following equivalence operators:
\begin{itemize}[itemsep=0pt]
    \item
    $\code{Equiv}_{C,I}(a,b)
    =
    C\bigl(I(a,b),I(b,a)\bigr)$.
    This is the generic equivalence obtained from the configured implication $I$ and conjunction $C$.
    \item
    $\code{EquivSimilarity}(a,b)=1-|a-b|^p$.
    This operator defines equivalence directly as a similarity score. It is reflexive and symmetric, and it satisfies the Boolean boundary conditions for $p>0$. For $p=1$, \code{EquivSimilarity} coincides with $\code{Equiv}_{\code{AndLuk},\code{ImpliesLuk}}$.
\end{itemize}

\paragraph{Standard fuzzy-logic families.}
Table~\ref{tab:standard-families} summarizes the main standard fuzzy-logic families implemented in \sltn{}. Each family specifies a coherent choice of conjunction, disjunction, and implication, typically obtained through standard duality or residuation constructions~\citep{badreddine_logic_2022}. These predefined families provide common configurations of the \sltn{} \code{Logic} object, while still allowing individual operators to be overridden when a different differentiability or modeling behavior is desired.

\begin{table}[h]
\centering
\resizebox{0.8\textwidth}{!}{
\begin{tabular}{lllll}
\toprule
Family & Conjunction & Disjunction & S-implication & R-implication \\
\midrule
G\"odel
& $\min(a,b)$
& $\max(a,b)$
& $\max(1-a,b)$
& $
\begin{cases}
1, & a \le b,\\
b, & \text{otherwise}
\end{cases}$ \\
Product/Goguen
& $ab$
& $a+b-ab$
& $1-a+ab$
& $
\begin{cases}
1, & a \le b,\\
b/a, & \text{otherwise}
\end{cases}$ \\
\L{}ukasiewicz
& $\max(a+b-1,0)$
& $\min(a+b,1)$
& $\min(1-a+b,1)$
& $\min(1-a+b,1)$ \\
\bottomrule
\end{tabular}}
\caption{Common fuzzy-logic operator families implemented in \sltn{}.}
\label{tab:standard-families}
\end{table}

\paragraph{Aggregation, quantifiers, and satisfaction.}
An \emph{aggregation operator} is a family of functions
\[
A_n:[0,1]^n \to [0,1],
\qquad n\ge 1,
\]
that is monotone in each argument and satisfies the boundary conditions
\[
A_n(0,\ldots,0)=0,
\qquad
A_n(1,\ldots,1)=1,
\]
together with the unary identity condition
\[
A_1(x)=x
\qquad
\text{for all } x\in[0,1].
\]
In \sltn{}, aggregation operators provide the common semantic mechanism for universal quantification, existential quantification, structural quantification, and satisfaction aggregation $\SatAgg$.

To accommodate guarded aggregation, each aggregator accepts inputs together with fuzzy mask values:
\[
A_n^{\mathrm{mask}}
:
[0,1]^n\times[0,1]^n
\to
[0,1].
\]
The evaluation of an aggregator on truth values $x = (x_i)$ and mask values $m=(m_i)$ is denoted:
\[
A_n^{\mathrm{mask}}(x;m).
\]
The unmasked case is recovered by setting $m_i=1$ for all $i$. Masks generalize ordinary aggregation by assigning an inclusion degree to each grounded instance. A Boolean mask $m_i\in\{0,1\}$ induces a \emph{hard} guarded aggregation, in which excluded instances are removed from the aggregation domain. A fuzzy mask $m_i\in[0,1]$ induces a \emph{soft} guarded aggregation, in which each instance contributes proportionally to its mask value. 

\noindent
\sltn{} implements the following aggregation operators:
\begin{itemize}[itemsep=0pt]
    \item $\code{AggregMin}(x)=\min_i x_i$. The parameter $\code{bottom\_k}$ returns the mean of the bottom-$k$ selected values. Supports Boolean masks only.
    \item $\code{AggregMax}(x)=\max_i x_i$. The parameter $\code{top\_k}$ returns the mean of the top-$k$ selected values. Supports Boolean masks only.
    \item
    \(
    \code{AggregGeometricMean}(x;m)
    =
    \exp\!\left(
    \frac{\sum_i m_i\log x_i}{\sum_i m_i}
    \right).
    \)
    This weighted geometric mean gives a multiplicative aggregation semantics and is sensitive to low truth values.
    \item
    \(
    \code{AggregPMean}(x;m)
    =
    \left(
    \frac{\sum_i m_i x_i^p}{\sum_i m_i}
    \right)^{1/p}
    \)
    with $p\in(0,+\infty)$. This operator computes the power mean of exponent $p$ \cite{bullen2013handbook}. More details on this operator can be found in the next paragraph.
    \item
    \(
    \code{AggregPMeanError}(x;m)
    =
    1-
    \left(
    \frac{\sum_i m_i(1-x_i)^p}{\sum_i m_i}
    \right)^{1/p}
    \)
    with $p\in(0,+\infty)$.
    This operator applies the power mean of exponent $p$ to the errors $1-x_i$.
\end{itemize}

\paragraph{Power-mean aggregation.}

\begin{figure}[htbp]
    \centering
    \resizebox{0.78\textwidth}{!}{
    \begin{tikzpicture}[
        font=\small,
        axis/.style={thick},
        tick/.style={thick},
        labelbox/.style={
            rounded corners=2pt,
            draw=black!30,
            fill=white,
            inner sep=3pt,
            align=center
        },
        endpoint/.style={
            circle,
            draw=black,
            fill=white,
            thick,
            minimum size=5mm,
            inner sep=0pt
        }
    ]

    \coordinate (Pzero) at (0,0);
    \coordinate (Pone)  at (5,0);
    \coordinate (Pinf)  at (10,0);

    \shade[left color=white, right color=black!35]
        (0,-0.15) rectangle (5,0.15);
    \shade[left color=black!35, right color=black!75]
        (5,-0.15) rectangle (10,0.15);

    \draw[axis] (0.,0) -- (10.,0);

    \foreach \x/\lab in {
        0/{p=0},
        2.5/{0<p<1},
        5/{p=1},
        7.5/{1<p<+\infty},
        10/{p=+\infty}
    }{
        \draw[tick] (\x,0.16) -- (\x,-0.16);
        \node[below=4pt] at (\x,-0.16) {$\lab$};
    }

    \draw[tick] (0.,0.2) -- (0.,0.6);
    \draw[tick] (10.,0.2) -- (10.,0.6);

    \node[endpoint] at (Pzero) {};
    \node[endpoint, fill=black!20] at (Pone) {};
    \node[endpoint, fill=black!80] at (Pinf) {};

    \node[labelbox, above=22pt] at (Pzero)
        {\code{AggregGeometricMean}};

    \node[labelbox, above=22pt] at (Pone)
        {\code{AggregPMean}};

    \node[labelbox, above=22pt] at (Pinf)
        {\code{AggregMax}};

    \node[align=center, below=25pt] at (1.25,0)
        {geometric-like\\ aggregation};

    \node[align=center, below=25pt] at (5,0)
        {uniform\\ averaging};

    \node[align=center, below=25pt] at (8.75,0)
        {max-like\\ aggregation};

    \draw [decorate, decoration={brace, amplitude=6pt}]
        (1,.5) -- (9, .5);

    \end{tikzpicture}
    }
    \caption{Spectrum of the power-mean aggregators across exponents $p$. The case $p=1$ gives arithmetic averaging; as $p\to0$ the operator approaches geometric aggregation, while as $p\to+\infty$ it approaches the maximum.}
    \label{fig:pmean_spectrum}
\end{figure}
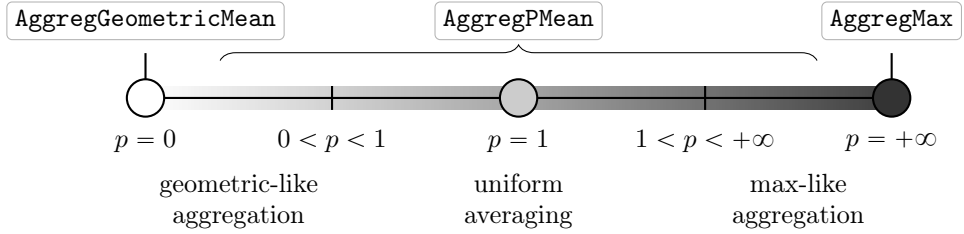

Let us consider the unmasked power-mean aggregator
\[
\code{AggregPMean}_p(x)
=
\left(
\frac{1}{n}\sum_{i=1}^n x_i^p
\right)^{1/p},
\qquad
p\in(0,+\infty),
\]
for truth values $x=(x_1,\ldots,x_n)$. It defines a continuous family of behaviours controlled by $p$. For positive inputs, the limit $p\to0$ is the geometric mean,
\[
\lim_{p\to0}
\code{AggregPMean}_p(x)
=
\left(\prod_{i=1}^n x_i\right)^{1/n} =\code{AggregGeometricMean}(x).
\] For $p=1$, \code{AggregPMean} is the arithmetic mean,
\[
\code{AggregPMean}_{p=1}(x)
=
\frac{1}{n}\sum_{i=1}^n x_i.
\]
As $p\to+\infty$, it approaches the maximum,
\[
\lim_{p\to+\infty}
\code{AggregPMean}_p(x)
=
\max_i x_i,
\]
corresponding to \code{AggregMax}. Thus, increasing $p$ shifts the aggregation from averaging behaviour toward maximum-like behaviour, giving greater influence to high truth values (see Figure \ref{fig:pmean_spectrum}).

\paragraph{Choice of aggregators.}
In crisp first-order logic, quantifiers reduce to extrema over the quantified
domain: existential quantification uses a maximum, whereas universal
quantification uses a minimum. A formula \(\exists x\,\varphi(x)\) is true if
at least one grounding of \(\varphi\) is true; conversely,
\(\forall x\,\varphi(x)\) is true only if every grounding is true. In
differentiable fuzzy reasoning, \sltn{} replaces these hard extrema with smooth
aggregation operators.

Existential quantification is naturally relaxed by \code{AggregPMean}. For
truth values \(x=(x_1,\ldots,x_n)\),
\[
\code{AggregPMean}_p(x)
=
\left(
\frac{1}{n}\sum_{i=1}^n x_i^p
\right)^{1/p},
\qquad p>0.
\]
As \(p\to+\infty\), this operator approaches the maximum:
\[
\lim_{p\to+\infty}
\code{AggregPMean}_p(x)
=
\max_i x_i.
\]
Thus, larger values of \(p\) make \code{AggregPMean} closer to the crisp
existential semantics, while finite values allow gradients to pass through
multiple grounded instances rather than only through the maximizer.

Universal quantification is naturally relaxed by the error-based dual
\code{AggregPMeanError}. With standard negation, this corresponds to the
double-negation transformation
\[
\forall z\,\varphi(z)
\equiv
\neg\exists z\,\neg\varphi(z),
\]
yielding
\[
\code{AggregPMeanError}_p(x)
=
1-
\left(
\frac{1}{n}\sum_{i=1}^n (1-x_i)^p
\right)^{1/p}.
\]
As \(p\to+\infty\), this operator approaches the minimum:
\[
\lim_{p\to+\infty}
\code{AggregPMeanError}_p(x)
=
\min_i x_i.
\]
Thus, \code{AggregPMeanError} is the corresponding smooth relaxation of
universal quantification.

The opposite regime \(p\to0\) does not recover a crisp quantifier. Instead, for
inputs away from the relevant boundary, it yields geometric aggregation:
\[
\lim_{p\to0}
\code{AggregPMean}_p(x)
=
\left(\prod_{i=1}^n x_i\right)^{1/n},
\]
and, dually,
\[
\lim_{p\to0}
\code{AggregPMeanError}_p(x)
=
1-
\left(\prod_{i=1}^n (1-x_i)\right)^{1/n}.
\]
This geometric-mean regime is less extremal than the max/min limits and tends
to distribute the learning signal over more grounded instances. It can be
useful in the early stages of training, or when one wants to avoid the
single-witness behavior of existential aggregation and the single-violation
behavior of universal aggregation. However, because geometric and power-mean
expressions involve products, powers, and logarithmic computations, they may be
sensitive to boundary values near \(0\) or \(1\). This motivates the
stabilization mechanisms described in Section~\ref{s:stability}.

In practice, \(p\) may also be scheduled during training, starting from a
smoother aggregation regime and gradually moving toward a more extremal
semantics.

\paragraph{Default behavior.}
When an \sltn{} \code{Logic} object is instantiated without explicit operator
choices, \sltn{} uses the stable-product configuration summarized in
Table~\ref{tab:default-logic}. This configuration follows the operator choices
commonly used in Logic Tensor Networks, with stabilization enabled by default
where appropriate. Stabilization mechanisms are discussed in
Section~\ref{s:gradient-analysis}.

\begin{table}[h]
\centering
{\renewcommand{\arraystretch}{1.25}
\begin{tabular}{lll}
\toprule
Logical role & Default \sltn{} class & Default parameters \\
\midrule
Negation $\neg$ & \code{NotStandard} & -- \\
Conjunction $\land$ & \code{AndProd} & \code{stable=True} \\
Disjunction $\lor$ & \code{OrProbSum} & \code{stable=True} \\
Implication $\rightarrow$ & \code{ImpliesGoguen} & \code{stable=True} \\
Equivalence $\leftrightarrow$ & \code{Equiv} & built from default $\land$ and $\rightarrow$ \\
Universal quantification $\forall$ & \code{AggregPMeanError} & \code{p=2, stable=True} \\
Existential quantification $\exists$ & \code{AggregPMean} & \code{p=2, stable=True} \\
Satisfaction aggregation $\SatAgg$ & \code{AggregPMeanError} & \code{p=2, stable=True} \\
\bottomrule
\end{tabular}}
\caption{Default stable-product configuration instantiated by \code{Logic()}.}
\label{tab:default-logic}
\end{table}

If the user supplies only part of a \code{Logic} configuration, the method
\code{with\_defaults()} completes the missing operators compositionally when
possible. In particular, if a conjunction and a negation are provided,
disjunction can be generated through \code{OrDual}, implication through the
S-implication construction \code{ImpliesDual}, and equivalence through mutual
implication via \code{Equiv}. Aggregators are completed using the defaults in
Table~\ref{tab:default-logic}, unless explicitly overridden.

\paragraph{Classical preset.}
For debugging and evaluation, \sltn{} provides a crisp Boolean-like preset through \code{Logic.classical()}. This preset substitutes hard, non-stable, and non-smooth operators for each logical role, as summarized in Table~\ref{tab:classical-logic}. On truth values in $\{0,1\}$, it recovers ordinary Boolean semantics exactly. The classical preset is intended for evaluation and diagnostic purposes rather than gradient-based learning.

\begin{table}[h]
\centering
{\renewcommand{\arraystretch}{1.25}
\begin{tabular}{ll}
\toprule
Logical role & Classical \sltn{} class \\
\midrule
Negation $\neg$ & \code{NotStandard} \\
Conjunction $\land$ & \code{AndMin} \\
Disjunction $\lor$ & \code{OrMax} \\
Implication $\rightarrow$ & \code{ImpliesGodel} \\
Universal quantification $\forall$ & \code{AggregMin} \\
Existential quantification $\exists$ & \code{AggregMax} \\
Satisfaction aggregation $\SatAgg$ & \code{AggregMin} \\
\bottomrule
\end{tabular}}
\caption{Classical debugging configuration provided by \code{Logic.classical()}.}
\label{tab:classical-logic}
\end{table}

\subsection{Gradient Pathologies}
\label{s:gradient-analysis}

In differentiable fuzzy reasoning, algebraic adequacy is not sufficient:
operators must also provide informative and numerically stable gradients.
Following prior analyses of fuzzy operators in differentiable settings
\citep{badreddine_logic_2022,van_krieken_analyzing_2020}, three gradient
pathologies are especially relevant:
\begin{itemize}[itemsep=0pt]
    \item
    \emph{Single-passing gradients} occur when only one input receives a
    non-zero derivative. This is typical of hard extrema such as
    \code{AndMin}, \code{AggregMin}, and \code{AggregMax}, where the
    gradient is propagated only through the currently selected extremal input.
    Such operators provide sparse learning signals.
    \item
    \emph{Vanishing gradients} occur when derivatives become close to zero over
    relevant regions of the domain. This is common for multiplicative
    operators, where the derivative with respect to one argument is scaled by
    the truth values of other arguments. As a result, poorly satisfied formulas
    may provide little learning signal.
    \item
    \emph{Exploding gradients} occur when derivatives become unbounded near
    the boundary of the truth interval. This may lead to unstable parameter
    updates or numerical failures, especially for operators involving powers,
    logarithms, or divisions near \(0\) or \(1\).
\end{itemize}

\begin{example}
The product t-norm illustrates vanishing gradients in multiplicative
conjunctions:
\[
\code{AndProd}(a,b)=ab.
\]
Its partial derivatives are
\[
\partial_a \code{AndProd}(a,b)=b,
\qquad
\partial_b \code{AndProd}(a,b)=a.
\]
Thus, when both inputs are close to zero, both derivatives are close to zero.
More generally, if one conjunct is weakly satisfied, the gradient with respect
to the other conjunct is attenuated. In nested product conjunctions, this
effect can compound across many conjuncts.

Power-mean and geometric-mean operators provide alternative gradient profiles.
For
\[
\code{AndPMean}_p(a,b)
=
\left(\frac{a^p+b^p}{2}\right)^{1/p},
\]
the derivative with respect to \(a\) is
\[
\partial_a \code{AndPMean}_p(a,b)
=
\frac{1}{2}a^{p-1}
\left(\frac{a^p+b^p}{2}\right)^{1/p-1}.
\]
As \(p\to0\), the operator approaches the geometric mean \(\sqrt{ab}\). This
can reduce the shrinkage caused by repeated products, but the power expression
may become numerically sensitive near the lower boundary. The geometric-mean aggregator illustrates boundary-induced instability:
\[
\code{AggregGeometricMean}(x_1,\ldots,x_n)
=
\exp\left(
\frac{1}{n}\sum_i\log x_i
\right).
\]
Its derivative is
\[
\partial_{x_i}
\code{AggregGeometricMean}(x_1,\ldots,x_n)
=
\frac{1}{n}
\frac{\code{AggregGeometricMean}(x_1,\ldots,x_n)}{x_i}.
\]
This derivative may become large as \(x_i\to0\), and the logarithm is undefined
at \(x_i=0\). Boundary stabilization is therefore required when such operators
are used in gradient-based learning.
\end{example}

\subsection{Stability Mechanisms}
\label{s:stability}

Several \sltn{} operators expose stabilization options to mitigate the
pathologies described above. These mechanisms address two complementary issues:
boundary singularities, which arise when operators are evaluated exactly at
\(0\) or \(1\), and non-smooth definitions, which arise from hard extrema,
case distinctions, or clamping.

\paragraph{Boundary projections.}
Operators with \code{stable=True} apply an affine projection before evaluating
the raw operator. Two projections are used:
\begin{align}
\pi_0(a) &= (1-\epsilon)a+\epsilon, \\
\pi_1(a) &= (1-\epsilon)a.
\end{align}
The projection \(\pi_0\) maps \([0,1]\) into \([\epsilon,1]\), moving values
away from the lower boundary. The projection \(\pi_1\) maps \([0,1]\) into
\([0,1-\epsilon]\), moving values away from the upper boundary. These
projections follow the stabilization strategy used in differentiable fuzzy
logics \citep{badreddine_logic_2022,carraro2024ltntorch}.

Operators involving products, powers, logarithms, or divisions typically use
\(\pi_0\), since their numerical difficulties often occur near \(0\). Operators
whose unstable regimes occur near \(1\) use \(\pi_1\). Stabilization changes
the exact numerical value of the operator near the boundary, but it improves
the reliability of gradient-based optimization.

\paragraph{Smooth approximations.}
Some fuzzy operators contain hard extrema or piecewise definitions. Examples
include the G\"odel and Goguen implications, as well as the \L{}ukasiewicz
conjunction:
\[
\code{AndLuk}(a,b)
=
\max(a+b-1,0).
\]
Such definitions are algebraically well motivated, but they may provide sparse
or discontinuous gradient information. When smooth variants are enabled, hard
branches or clamps are replaced by differentiable approximations, such as
softplus- or squareplus-based relaxations~\citep{squareplus}. These
approximations introduce a small gradient signal near transition regions and
can improve optimization stability, at the cost of no longer matching the
exact piecewise operator everywhere.

\end{document}